\documentclass{article}

\PassOptionsToPackage{numbers,sort&compress}{natbib}
\usepackage[preprint]{neurips_2024}

\usepackage[utf8]{inputenc} 
\usepackage[T1]{fontenc}    
\usepackage{hyperref}       
\usepackage{url}            
\usepackage{booktabs}       
\usepackage{amsfonts}       
\usepackage{subfig}         
\usepackage{nicefrac}       
\usepackage{microtype}      
\usepackage{xcolor}         

\usepackage{enumitem}       
\usepackage{amsmath}
\usepackage{amstext}        
\usepackage{doi}
\usepackage{graphicx}       
\usepackage{float}

\usepackage{multirow,setspace,caption}
\usepackage{tikz}
\usepackage{array}

\usepackage{rotating}

\usepackage{booktabs, array, siunitx}
\newcolumntype{L}[1]{>{\raggedright\arraybackslash}p{#1}}
\newcolumntype{R}[1]{>{\raggedleft\arraybackslash}p{#1}}

\title{3D Underwater Path Planning via Generative Flow Field Surrogates}

\author{%
  Zachary Cooper-Baldock\\
  Flinders University\\
  \And 
  Paulo E. Santos \\
  Flinders University \\
  PrioriAnalytica \\
  \And
  Russell S.A. Brinkworth \\
  Flinders University \\
  \And
  Karl Sammut \\
  Flinders University \\
}

\begin{document}

\maketitle

\begin{abstract}
Autonomous underwater vehicle (AUV) launch and recovery (LAR) into the hull of an advancing host platform requires traversal of a complex, three-dimensional propeller wake whose hydrodynamic structure cannot be characterised by a uniform current model. High-fidelity Reynolds-Averaged Navier-Stokes (RANS) Computational Fluid Dynamics (CFD) simulations resolve this structure with sufficient accuracy for path planning, but their computational cost renders them impractical for onboard use. We address this gap by integrating two conditional generative adversarial network (cGAN) architectures --- a regularised PatchGAN and a 2D3DGAN with self-attention --- as drop-in replacements for RANS CFD data within a three-dimensional, energy-weighted A* path planning framework. Both generators are driven by a hierarchical pipeline that synthesises full $128^3$ voxel flow field volumes from scalar operating condition inputs alone, with end-to-end inference times of approximately 28--146~$\mu$s, compared to hours for a single RANS computation. We benchmark all four environmental knowledge levels: uniform current, ground-truth CFD, PatchGAN, and 2D3DGAN~SA across 19,800 independently generated trajectories spanning 550 distinct flow conditions. Full CFD wake knowledge reduces energy expenditure by 5.7--12.5\% and high-velocity wake-core encounters by up to 77.8\% relative to uniform-current planning, with both benefits scaling with operating severity. The cGAN surrogates recover approximately 45--60\% of the CFD energy benefit and high-velocity cell avoidance benefit while operating at inference speeds compatible with edge device use. These results provide the first systematic quantification of the downstream path planning value of cGAN-predicted hydrodynamic fields in a three-dimensional maritime robotics application.
\end{abstract}


\section{Introduction}
\label{sec:Introduction}

Autonomous Underwater Vehicles (AUVs) are increasingly deployed in complex, multi-vehicle operational scenarios where their proximity to other platforms is unavoidable. Among the most demanding of these is underwater launch and recovery (LAR). In this action, a smaller vehicle must navigate toward or from the stern of a larger advancing host platform, such as an extra-large uncrewed underwater vehicle (XLUUV), into a payload bay recessed within the host's hull \cite{ZacIEEE, ZacSubSTEC, ZacPlanner}. This manoeuvre requires the recovering vehicle to traverse the propeller race and turbulent wake of the host platform, exposing it to complex, three-dimensional (3D) hydrodynamic forces that drive significant energy expenditure and pose real risks to control authority and navigational stability. The severity of the hydrodynamic interaction scales with proximity. At close range, localised wake structures including velocity fluctuations, shear layers and tip vortices shed by the XLUUV propeller and body geometry become dominant environmental loads on the vehicle, far exceeding the effects of the ambient current alone \cite{ZacIEEE}.

Effective path planning for such manoeuvres requires an accurate, timely and effective representation of the 3D flow field environment. However, a fundamental tension exists between the fidelity of available environmental models and the computational constraints of onboard, real-time decision-making. High-fidelity Reynolds-Averaged Navier-Stokes (RANS) Computational Fluid Dynamics (CFD) simulations are capable of resolving the full 3D wake structure with sufficient accuracy for path planning, but their computational cost demands substantial hardware resources and simulation times, far exceeding the mission duration \cite{ZacPlanner}. This renders them impractical for onboard use. As a result, existing path planners for underwater vehicles predominantly rely on simplified environmental representations. These can take the form of uniform or depth-averaged current fields, 2D flow approximations, or models that reduce the flow to a scalar obstacle-avoidance problem. While computationally tractable, these approaches fail to capture the localised wake structures that most significantly affect AUV navigation during close-proximity operations. This creates a critical gap between planning optimality and practical deployability.

The path planning literature for autonomous underwater systems has advanced considerably in its treatment of global ocean currents, geometric obstacle avoidance, and multi-objective optimisation. Common algorithmic frameworks, including node-based methods such as A*, sampling based approaches such as RRT, and bio-inspired methods such as Ant Colony Optimisation have each been adapted to incorporate environmental flow effects. However, with few exceptions, these works are confined to 2D domains or use simplified current models that extend 2D data linearly into 3D space, neglecting the full volumetric structure of vehicle wakes. Studies addressing turbulent 3D environments, principally in the context of UAV flight \cite{Pensado2024} have characteristically responded by routing vehicles entirely around turbulent regions: an approach that is geometrically and operationally infeasible for AUV LAR, where the wake structure must be traversed and cannot be circumvented. The consequence is that, to date, no systematic quantification has been made of what path planning performance is lost by ignoring 3D wake structures in this application, nor of the degree to which any computationally tractable surrogate model can recover those losses.

Generative Adversarial Networks (GANs) and specifically their conditional variants (cGANs) have emerged as a compelling class of data-driven surrogate model for fluid dynamics applications. When trained on high fidelity CFD data, cGANs can learn to generate volumetric flow field approximations from sparse conditional inputs \cite{ZacLoss, Zac_PhD_Thesis} such as planar velocity slices and scalar operating condition metadata, at inference speeds that are orders of magnitude faster than direct numerical simulation \cite{Zac_PhD_Thesis}. Recent advances in 3D flow field reconstruction from 2D observations have demonstrated that such models can achieve high fidelity reconstruction at resolutions relevant to engineering applications \cite{Existing3DSota}. In the context of AUV path planning, this raises a directly practical question: can a cGAN-predicted flow field serve as an adequate substitute for ground-truth CFD data within a planning loop, and if so, how much of a performance benefit of full hydrodynamic knowledge does it retain?

In prior work \cite{ZacPlanner}, the authors established a baseline comparison between a fully CFD-informed A* path planner and a current-only (naive) planner for the AUV LAR scenario, demonstrating that incorporating the RANS-computed wake structures reduces energy expenditure by up to 12.09\% and reduces encounters with high-velocity flow regions by up to 77.83\%. That work validated the core hypothesis that detailed hydrodynamic wake knowledge is operationally significant, but was limited to the idealised case of perfect CFD information about the wake, vs only knowledge of the current, leaving open the central question of whether any practically deployable surrogate model could bridge this gap.

The present work directly addresses that question. We extend the prior benchmark to a four-way comparison by integrating two cGAN architectures: a regularised PatchGAN and a 2D3DGAN with self-attention (2D3DGAN SA),  as drop in replacements for RANS CFD data within the planning framework. These models, trained on a dataset of 550 distinct RANS flow conditions extracted from \cite{WAKESET}. The flow conditions span operating speeds from 0.10 to 5.00 m/s and heading angles from 0$\deg$ to 50$\deg$, with the cGAN architectures used to generate full $128^3$ volumetric flow fields from conditioning scalar metadata via a hierarchical 1D-to-2D-to-3D prediction pipeline. By evaluating all four planner variants: current-only, RANS CFD, PatchGAN and 2D3DGAN SA - across 19,800 independently generated trajectories under matched conditions, we produce the first systematic quantification of the downstream path planning value of cGAN-predicted hydrodynamic fields in a 3D maritime robotics application.

\subsection{Contributions}
\label{sec:Contributions}

The specific contributions of this work are as follows:

\begin{enumerate}
    \item We present the first systematic, multi-fidelity benchmark of 3D path planning performance for AUV launch and recovery, comparing four environmental knowledge levels: uniform current (naive), RANS CFD, PatchGAN-predicted fields and 2D3DGAN SA-predicted fields, evaluated across 19,800 trajectories spanning 550 distinct flow conditions. This provides a quantitative characterisation of the performance cost of environmental information and the degree to which each surrogate level recovers the benefits of full CFD knowledge.
    
    \item We introduce and validate a hierarchical cGAN prediction pipeline for real-time 3D flow field generation, in which a 2D cGAN produces the conditional planar inputs required by a 3D volumetric generator from minimal onboard sensor data (vehicle speed and heading). The pipeline is demonstrated to produce flow field predictions at inference speeds compatible with onboard deployment, in contrast to the impractical compute requirements of RANS CFD simulation.

    \item We develop a holistic five-metric evaluation for path optimality encompassing energy expenditure, path length, encounters with high velocity regions, turbulent cell transitions and compute time and demonstrate through its application that optimising for energy efficiency introduces a non-trivial trade-off with turbulence exposure, a finding with direct implications for AUV control system design in wake-dominated environments.
    
\end{enumerate}

The remainder of this paper is organised as follows. Section \ref{sec:Related_Work} reviews the relevant path planning literature and positions this work relative to existing 3D flow-informed planning methods. Section \ref{sec:Hydrodynamic_Domain} describes the computational domain, the RANS CFD dataset and the LAR scenario configuration. Section \ref{sec:Path_Planning_Framework} presents the energy-weighted A* planning framework and experimental design for trajectory generation. Section \ref{sec:GAN_Pipeline} details the cGAN architectures and the hierarchical prediction pipeline used to generate the four environmental models. Section \ref{sec:Evaluation_Metrics} defines the path quality evaluation metrics. Section \ref{sec:Results} presents and analyses the comparative results across all planners and flow conditions. Section \ref{sec:Discussion} discusses the broader implications of the findings, addresses the principal limitations of the study, and identifies directions for future work. Section \ref{sec:Conclusion} concludes and indicates directions of future work. 

\section{Related Work}
\label{sec:Related_Work}

This work sits at the intersection of three bodies of literature: AUV hydrodynamics and launch and recovery operations, path planning for autonomous underwater vehicles in flow-dominated environments, and data-driven surrogate modelling of fluid dynamics using generative adversarial networks. The following review is structured accordingly, culminating in a synthesis that identifies the specific gap addressed by this work.

\subsection{AUV Hydrodynamics in Multi-Vehicle Operations and Launch and Recovery}
\label{subsec:AUVHydrodynamicsReview}

Close-proximity hydrodynamic interactions between underwater vehicles present a substantively different operational environment than that of an isolated vehicle in open water. As the separation between platforms reduces, the wake, boundary layer, and propeller race of the lead vehicle increasingly dominate the flow experienced by the trailing vehicle creating time-varying forces and moments that can exceed the vehicle's control authority if not accounted for during navigation. These effects are well documented in the surface ship domain for towing and replenishment operations, and more recently for fully submerged multi-vehicle configurations \cite{ZacIEEE}.

The hydrodynamic feasibility and complexity of the specific LAR configuration considered in this work: a smaller AUV approaching and entering a payload bay within the hull of a larger XLUUV, has been established in dedicated CFD and experimental studies \cite{ZacIEEE, ZacSubSTEC, ZacPlanner}. These analysis demonstrated that the approaching vehicle encounters a strongly three-dimensional flow environment that is dominated by the host vehicle's propeller race and trailing wake. Crucially, the magnitude and spatial distribution of these hydrodynamic loads are highly sensitive to the relative heading and speed of the host vehicle, and the loads cannot be adequately approximated by a uniform or depth-averaged model. Wake velocity deficits, tip vortices, and propeller-induced swirl all contribute to a spatially complex velocity field that varies significantly between flow conditions. This operational evidence provides the physical motivation for requiring a full volumetric representation of the flow field in any planning framework that seeks to minimise energy expenditure or reduce control disturbances during the approach manoeuvre. 

The significance of propeller wake effects on AUV control during close-proximity operations has been examined in the context of underwater docking and formation operations more broadly \cite{ZacIEEE, KarlSuggested, Yi2023}. These studies collectively reinforce that the six-degree-of-freedom dynamics of an approaching AUV are substantially modified by the wake of the host, but that the spatial distribution of the resulting hydrodynamic forces cannot be inferred from a simple scalar current model. Despite this well-established evidence base, the majority of path planning frameworks applied to underwater vehicle operations have not integrated this knowledge, a gap that this work directly addresses.

\subsection{Path Planning for Autonomous Underwater Vehicles}
\label{subsec:PathPlanningReview}

Path planning for AUVs is a mature field, with extensive literature addressing the optimisation of trajectories with respect to travel time, energy consumption, and collision avoidance \cite{X1, X3, X4, X5}. The environmental representation used within the planner, acting as the model of the forces and obstacles the vehicle will encounter, a critical design choice that fundamentally determines the quality and operational validity of the resulting trajectories. The following review examines how this environmental representation has evolved, and where its current limitations lie. 

\subsubsection{Node-Based and Sampling-Based Planning Algorithms}
Node-based search algorithms, and particularly A*, remain widely used in AUV path planning due to their optimality guarantees when combined with an admissible heuristic \cite{Hart1968}. Applications in 2D domains have demonstrated the ability to account for spatially variable ocean currents in the cost function, generating energy-aware trajectories that exploit or avoid flow regions \cite{Garau2005, Petres2007, Kruger2007, Li2016, Zadeh2018}. Sampling based methods, such as the Rapidly-exploring Random Tree (RRT) and its optimal variant, RRT* \cite{LaValle2006, RandomTrees}, offer probabilistic completeness. This is well suited to high-dimensional configuration spaces with complex obstacles. Bio-inspired methods such as Ant Colony Optimisation \cite{AntColony} and evolutionary algorithms \cite{Alvarez2004} have also been applied to current-aware trajectory optimisation. Across these algorithmic families, however, the common limitation is that the environmental representations are predominantly 2D, and where 3D formulations have developed, the flow is typically a simplified or linearly extended version of 2D current data that omits the complex volumetric structure of vehicle wakes \cite{Kruger2007, Li2016, Zadeh2018, KarlSuggested, Yi2023, Pensado2024, Pfeiffer2017}.

\subsubsection{3D Path Planning with Flow Effects}

Several works have extended AUV path planning to genuinely 3D formulations that incorporate current effects. A* has been applied to estuary current fields in 3D \cite{Kruger2007}, and optimisation-based approaches have incorporated 3D current data for time- and energy-optimal routing \cite{Li2016}. Differential evolution has been applied to 3D trajectory planning that accounts for ocean currents and volumetric obstacles \cite{Zadeh2018}, and constrained optimisation approaches have incorporated multi-objective criteria including travel distance and current effects in three dimensions \cite{Yi2023}. Each of these contributions represents a meaningful advance over 2D methods, but in doing so, assumes a simplified current field that varies smoothly in space, without accounting for the concentrated, high-gradient wake structures generated by a nearby vehicle.

The most relevant precedent for the present work is a study of A*-based navigation in turbulent 3D aerial domains \cite{Pensado2024}, in which turbulent regions were identified and routed around entirely. While effective for UAV operations where the wake structure is spatially limited and avoidance is geometrically feasible, this approach cannot be transferred to the AUV LAR scenario. In this context, the wake of the host platform, including its propeller race, occupies the region through which the recovering vehicle is required to pass, and its coherent structures span length scales commensurate with the vehicle itself. Complete avoidance is therefore neither geometrically possible nor operationally meaningful. What is required instead is an informed traversal strategy. A planner that understands the 3D structure of the wake, well enough to select the path of least resistance through it. The prior publication \cite{ZacPlanner} on which this work builds, established that such a strategy is both feasible and operationally significant when using ground truth RANS CFD data. The present work extends this benchmark to include data-driven surrogate fields as the environmental input. 

\subsubsection{Machine Learning in Path Planning}
Neural network methods have been increasingly applied to path planning tasks to capitalise on their speed advantages over iterative search \cite{Tai2017, Pfeiffer2017}. Feed-forward networks and convolutional architectures trained via imitation learning or reinforcement learning can produce trajectory proposals at inference speeds several orders of magnitude faster than classical planners, enabling reactive re-planning on resource-constrained hardware. These approaches have demonstrated success in obstacle avoidance tasks in 2D grid environments and have been extended to 3D spaces for aerial robotics applications \cite{Tai2017, Pfeiffer2017}. However, these methods have not been applied to navigation within complex, fully characterised 3D hydrodynamic fields. Their environmental representations typically treat obstacles as binary occupancy grids, rather than as continuous velocity fields from which a cost surface for energy aware-planning can be derived. The integration of ML-based planning with a high-fidelity hydrodynamic model, which is the contribution of the present work, therefore remains an open problem.

\subsection{Data-Driven Surrogate Modelling of Fluid Dynamics}
\label{subsec:SurrogateModellingReview}
The computational cost of high-fidelity CFD simulation has driven substantial interest in data-driven surrogate models that can approximate the outputs of such simulations at a fraction of the computational cost \cite{SCA24}. The following review covers the principal classes of ML surrogate relevant to this work: physics-informed neural networks, CNN-based regression surrogates, and generative adversarial networks applied to flow field reconstruction.

\subsubsection{Physics-Informed Neural Networks and Regression Surrogates}
Physics-Informed Neural Networks (PINNs) \cite{PINN_OG} embed the governing partial differential equations of fluid mechanics directly into the training objective, penalising violations of mass and momentum conservation during optimisation. PINNs can operate with limited labelled training data, making them attractive when CFD datasets are expensive to generate. However, their convergence behaviour is well documented to be problematic for high Reynolds number turbulent flows \cite{PINN_Challenge1, PINN_Challenge2, PINN_worsethanexisting}. The stiffness of the Navier-Stokes equations and the broad spectrum of length and time scales in highly turbulent flow make the PDE residual a difficult loss surface to optimise. For the flows within this work, characteristic of vehicle-scale operations at several meters per second, PINNs have not been demonstrated to match the reconstruction fidelity that can be achieved by other data-driven architectures.

Convolutional neural networks (CNN) regression surrogates directly map from a set of input conditions to flow field outputs, bypassing the PDE residual entirely \cite{Cambridge3D_CNN}. These architectures have achieved high reconstruction accuracy for laminar and mildly turbulent flows in 2D, demonstrating that the mapping from boundary conditions or operating parameters to velocity fields can be effectively learned from data. Extensions to 3D domains have been demonstrated using 3D convolutional architectures \cite{Cambridge3D_CNN}, but the absence of adversarial training means that sharp flow features like vortex cores, boundary layer edges and wake shear layers are often smoothed during the regression process, reducing the spatial fidelity of the predicted field precisely in those regions that are most energetically significant for path planning.

\subsubsection{Generative Adversarial Networks for Flow Field Reconstruction}
Generative Adversarial Networks and their conditional variants (cGANs) \cite{cgans19, Mirza2014cGAN} introduce a discriminator network that is trained adversarially alongside the generator, providing a learned perceptual loss that encourages the generator to reproduce the statistical distribution of the training data rather than merely minimising pixel-wise error. For flow field applications, this adversarial objective has been shown to substantially improve the fidelity of reconstructed features, producing fields that are both quantitatively accurate and physically coherent in their spatial structure \cite{Yang2019PINNcGAN, Mao2020AirfoilcGAN, Ren2021FlowControlcGAN}.

The application of cGANs to 2D flow field super-resolution and reconstruction has been demonstrated for a range of canonical flows including turbulent channel flow and cylinder wake configurations and some AUV applications \cite{ZacLoss}. Architectures incorporating self-attention mechanisms \cite{zhang2019self} have been shown to improve the reconstruction of spatially extended structures by allowing any spatial location in the generated field to directly attend to any other, circumventing the local receptive field limitation of convolutional layers. PatchGAN discriminators \cite{isola2017image}, which evaluate the realism and coherence of overlapping local patches of the output rather than the entire field as a single scalar, provide a particularly effective training signal for flow fields by enforcing local structural fidelity across the full spatial domain of the output. 

The reconstruction of full 3D volumetric flow fields from sparse 2D observations is a significantly more challenging task, and the literature is correspondingly less mature. The foundational work of Yousif et al. \cite{Yousif_2021} demonstrated the reconstruction of 3D turbulent velocity fields at $48^3$ voxel resolution from two orthogonal 2D planar slices, using a cGAN architecture (2D3DGAN) that encodes the input slices through 2D CNNs before lifting the resulting features to a 3D volume for up-sampling. This work established the feasibility of the 2D-to-3D pipeline for volumetric flow reconstruction via adversarial training. This work was used as the architectural starting point for the 3D generators used in the present work which operate at $128^3$ resolutions, a nearly 19x increase in voxel count, and incorporate additional conditioning mechanisms including FiLM layers \cite{FiLMLayers}, cross-attention between the 2D input features and the 3D volume \cite{Existing3DSota} and a custom gradient-weighted MSE loss function \cite{ZacLoss} designed to preserve the sharp hydrodynamic gradients at wake boundaries and shear layers. Similar CNN-based reconstruction approaches at lower resolution have been demonstrated in \cite{Cambridge3D_CNN} and \cite{CFD_GAN_Buildings}, but without the adversarial training component and the associated gains in sharp-feature fidelity. 

\subsubsection{Downstream Task Evaluation of Flow Field Surrogates}
An important and under-explored question in the ML surrogate literature concerns the relationship between standard reconstruction metrics such as MSE, PSNR, SSIM, and FID, and the performance on downstream engineering or robotics tasks for which the surrogate is designed. High reconstruction accuracy, as measured by these metrics, is a necessary but not sufficient condition for useful downstream performance. A surrogate that accurately reconstructs the global flow structure but introduces local errors in energetically significant regions may perform poorly as a planning environment even when its reconstruction score is high. Conversely, a computationally lightweight model with modest reconstruction fidelity may nevertheless preserve the planning-relevant features of the flow field, such as the spatial structures and velocity magnitude of the wake core in a sufficient manner to recover the majority of the planning benefit.

This distinction between reconstruction fidelity and downstream task performance has received limited systematic attention in the fluid dynamics literature. Related work in aerodynamic shape optimisation \cite{Lee2020AirfoilGAN} and structural mechanics \cite{VAE-GANs} has noted that surrogate models evaluated solely on prediction accuracy can produce misleading conclusions about their utility in design loops. For path planning applications specifically, no prior work has systematically quantified how much of the planning benefit achievable with ground-truth CFD field is retained when the planner instead operates on a cGAN predicted field. This evaluation is the central contribution of the present work. 

This review identifies a clear and specific gap in the literature. AUV path planning has been extensively studied for environments characterised by global ocean currents and geometric obstacles, but the full volumetric wake structure of a host vehicle has not been incorporated into a 3D planning framework, nor has its operational significance been quantified relative to simpler environmental models. Separately, cGANs have been demonstrated as effective surrogates for 3D flow field reconstruction, but their utility as drop-in replacements for RANS CFD data within an engineering task and specifically, within a path planning loop, has not been evaluated. 

The present work addresses both gaps simultaneously. We embed two cGAN architectures: a lightweight regularised PatchGAN and a higher fidelity 2D3DGAN with self-attention as environmental models within a 3D A* path planning framework, and benchmarking the resulting planners against a current only baseline and a ground truth CFD planner across 19,800 trajectories. In doing so, we produce the first systematic quantification of: (i) the path planning value of 3D wake information in an AUV LAR scenario; and (ii) the fraction of that value retained by cGAN-predicted flow fields relative to the ground truth. This benchmark directly addresses the broader question of when and how cGAN surrogates are sufficient for demanding downstream robotics tasks.

\section{Hydrodynamic Domain}
\label{sec:Hydrodynamic_Domain}

This section describes the physical scenario, computational geometry, fluid properties, CFD methodology, and dataset structure that underpin all planning experiments in this work. The domain is used both as the source of high-fidelity ground-truth flow fields against which the cGAN surrogates are benchmarked, and as the physical environment within which all path planning trajectories are generated and evaluated.

\subsection{Launch and Recovery Scenario}

The operational scenario considered in this work is the underwater launch and recovery (LAR) of a small AUV into the payload bay of a larger, steadily advancing extra-large uncrewed underwater vehicle (XLUUV). During the approach, the AUV navigates upstream from a starting position aft of the XLUUV's stern, transits through the host vehicle's propeller race and trailing wake, and enters a bottom-mounted payload bay recessed within the XLUUV's hull. The hydrodynamic feasibility and complexity of this specific manoeuvre have been established in dedicated prior analyses \cite{ZacIEEE, ZacSubSTEC}, which demonstrated that the approaching vehicle is subjected to strong three-dimensional wake structures including propeller-induced swirl, tip vortices, and velocity deficits that cannot be characterised by a uniform current model and vary substantially with the host vehicle's speed and heading angle.

The domain is modelled as a pseudo-static environment: for each planned trajectory, the XLUUV's velocity and heading are held constant, and the flow field is treated as a steady-state realisation of that operating condition. This quasi-steady assumption is physically motivated by the separation of timescales inherent in the LAR manoeuvre. RANS simulations of the type used to generate the dataset converge to a steady mean flow solution that captures the time-averaged wake structure, including the propeller race and boundary layer, which are the dominant navigational features during the approach. The AUV approach duration for a single trajectory is short relative to the timescale over which these mean structures evolve, and the XLUUV is assumed to maintain constant speed and heading throughout. The pseudo-static treatment is therefore consistent with the pseudo-transient RANS methodology and is a widely adopted assumption in prior path planning studies for current-dominated environments \cite{ZacPlanner, Pensado2024}. The limitations this places on the applicability of the results to unsteady and manoeuvring scenarios are acknowledged and discussed in Section \ref{sec:DiscussionLimitations}.

\subsection{XLUUV and AUV Geometry}
\label{sec:VehicleGeometry}

The XLUUV model used in this study is a generalised reference geometry designed to represent the common structural features of existing large XLUUV platforms including the Boeing Orca and similar vehicles, while avoiding overfitting to the specific surface details of any single design \cite{WAKESET}. The geometry, shown in Table 1, comprises a 22 m length hull with a cross-section of 2.2 m (beam) × 2.7 m (depth), a rounded nose, and flattened sides with curved edges to limit crossflow separation. A bottom-mounted payload bay of dimensions 1.40 m (width) × 2.20 m (depth) × 6.00 m (length) is located centrally along the hull, with its centre 7 m from the bow. The bay is accessible from below, consistent with the payload bay LAR configuration assessed in \cite{ZacIEEE, ZacSubSTEC}. The XLUUV is equipped with an actuator disk model representing the propulsive effect of its propeller, with the disk parametrised to match the thrust required to overcome drag at each simulated operating speed.

\begin{table}[!ht]
\begin{center}
\small
                    \begin{tabular}{L{3.5cm} L{3.0cm} L{6.0cm}}
                    \toprule \toprule
                    \textbf{Component}     & \textbf{Value $[Unit]$} & \textbf{Design Basis} \\ \toprule \toprule
                    Hull length          & 22.0 [\textit{m}] & Representative XLUUV class; within range of existing designs \cite{ZacIEEE}  \\ \hline
                    Beam $\times$ depth  & 2.2 $\times$ 2.7 [\textit{m}] & Typical XLUUV cross-section diameter range 1.5 - 2.0 m \cite{2_1} \\ \hline
                    Payload bay (W $\times$ D $\times$ L) & 1.40  $\times$ 2.20  $\times$ 6.00 [\textit{m}] & Aspect ratio matched to Boeing Orca: bottom-centre of hull \cite{11, 2_1} \\ \hline
                    Payload bay centre (from bow) & 7.0 [\textit{m}] & 3m forward of stern slope onset \\ \hline
                    Operating depth      & 100 [\textit{m}] & Representative of AUV/XLUUV operating depth \cite{ZacIEEE} \\ \hline
                    Computational domain & 155 $\times$ 155 $\times$ 155 [\textit{m}] & Centred on XLUUV; wake captured to 8L downstream  \\ \toprule \toprule
                    \end{tabular}
                    \caption{XLUUV and domain geometry summary}
                    \label{tab:XLUUVSummary}
                    \end{center}
                \end{table}

The AUV is not explicitly geometrically represented in the RANS simulations used to generate the flow field dataset; the wake and boundary layer of the XLUUV are computed in isolation, consistent with the assumption that the hydrodynamic perturbation induced by the much smaller approaching vehicle on the host's flow field is negligible at the path-planning stage. The AUV is modelled as a rigid body for the purposes of the drag-based energy cost function in the path planner, with its physical particulars determined from prior experimental and numerical analysis of the vehicle class \cite{ZacIEEE, WAKESET} 

\subsection{RANS CFD Methodology}
\subsubsection{Governing Equations and Turbulence Model}

All flow field data used in this study were generated using the Reynolds-Averaged Navier–Stokes (RANS) equations, solved using ANSYS Fluent 2023R1. The incompressible, isothermal RANS equations for mean Cartesian velocity ($\overline{u_{i,j}}$) and mean pressure ($P$) are given by Equation \ref{eqn:RANS} \cite{56}.

\begin{equation}
    \label{eqn:RANS}
    \rho \frac{\partial \overline{U_{i}}}{\partial t} + \rho \frac{\partial \overline{U_{i} U_{j}}}{\partial x_{j}} = - \frac{\partial P}{\partial x_{i}} + \frac{\partial}{\partial x_{j}} \left \{ \mu \left ( \frac{\partial \overline{U_{i}}}{\partial x_{j}} + \frac{\partial \overline{U_{j}}}{\partial x_{i}} \right ) \right \} - \rho \frac{\partial \overline{u'_{i} u'_{j}}}{\partial x_{j}} + f_{i}
\end{equation}

The realisable k-$\epsilon$ turbulence model \cite{57, 58, 59} was selected for its established suitability for high-Reynolds-number turbulent flows and accurate pressure drag prediction, with reported agreement of approximately 3\% with experimental drag data for bodies of similar form \cite{60}. The near-wall region was modelled using standard wall functions \cite{60}, targeting a dimensionless wall distance of 30 < $y^+$ < 300. Post-simulation verification confirmed area-averaged y+ values of 38.6 on the AUV and 99.3 on the XLUUV, both within the valid range. The flow domain was defined as a rectangular prism with the XLUUV centred within it; the inlet was placed 3L (66 m) ahead of the bow and the outlet 8L (176 m) astern, with open pressure-driven boundaries on the remaining four faces, consistent with validated AUV simulation practice \cite{ZacIEEE}.

\subsubsection{Propeller Model}

The thrust effect of the XLUUV's propeller was represented using an actuator disk (virtual blade) model, parametrised to balance propulsive thrust against hydrodynamic drag at each simulated speed. The disk diameter was set to 1.25 m. The pressure jump across the disk was varied as a quadratic function of domain inflow speed \cite{61, 64}, ensuring that the propulsive thrust matched the computed drag at each operating condition throughout the speed range 0.10–5.00 m/s. The propeller tip speed was scaled linearly with domain speed, maintaining a constant advance ratio of J = 0.85. Modelling the propeller via an actuator disk avoids the geometric and temporal complexity of full blade-resolved simulation while capturing the dominant propulsion-induced flow structures principally the axial velocity deficit and the swirl in the propeller race that are the most significant sources of navigational perturbation during LAR \cite{ZacIEEE, ZacSubSTEC}.

\subsubsection{Fluid Properties}

The fluid properties were set to represent seawater at a representative AUV operating depth of 100 m and a temperature of 21.8 °C, based on BRAN2020 ocean analysis data for the Australasian region \cite{49}. The key properties used are summarised in Table \ref{tab:WaterProps}. These values were incorporated directly into the ANSYS Fluent solver.

\begin{table}[!ht]
\begin{center}
\small
                    \begin{tabular}{lll}
                    \toprule \toprule
                    \textbf{Parameter}     & \textbf{Value $[Unit]$} & \textbf{Determined via} \\ \toprule \toprule
                    Simulation depth $(d)$ & 100 $[m]$ &  Selection \\ \hline
                    Temperature $(T)$      & 21.8 $[^{\circ}C]$ & Calculation \cite{50} \\ \hline
                    Pressure $(P)$         & 1.0057 $[MPa]$ & Calculation \cite{50} \\ \hline
                    Molar mass $(M)$       & 18.434 $[kg/kmol]$ & Calculation \cite{51} \\ \hline
                    Density $(\rho)$       & 1025.1627$[kg/m^{3}]$ & Calculation \cite{52} \\ \hline
                    Specific heat capacity $(c)$ & 4003.3 $[J/Kg.KJ]$ &  Calculation \cite{53} \\ \hline
                    Thermal conductivity $(k)$   & 0.58981 $[W/mK]$ & Calculation \cite{53}  \\ \hline
                    Dynamic viscosity $(\mu)$    & 0.0010314 $[kg/m/s]$ & Calculation \cite{54} \\ \hline
                    Specific enthalpy $(h)$      & 86.961 $[kJ/kg]$ & Calculation \cite{52} \\ \hline
                    Thermal expansion coeff. $(CTE)$ & 271.4 $[K^{-1}]$ & Calculation \cite{55}  \\ \toprule \toprule
                    \end{tabular}
                    \caption{Physical and chemical properties of the seawater model used in the initial computational fluid dynamics analysis.  Values were assimilated and calculated from BRAN2020 analysis data for the Australasian region at a fixed depth of 100 meters. }
                    \label{tab:WaterProps}
                    \end{center}
                \end{table}

\subsubsection{Mesh and Convergence}

A hybrid unstructured mesh was employed, comprising poly-hexcore cells in the domain interior with localised refinement bodies of influence (BOIs) around the propeller disk, propeller wake, and payload bay, and an inflation layer at all solid surfaces. A dedicated mesh and iteration convergence study was conducted specifically for the generalised XLUUV geometry and expanded parameter space, assessing convergence of area-weighted facet-averaged dynamic pressure, total pressure, turbulent kinetic energy, wall shear stress, and vorticity across seven mesh resolutions ranging from 1.38 × $10^6$ to 38.3 × $10^6$ cells. Convergence in all monitored quantities was achieved at 10–15 × $10^6$ cells; a mesh size of 14 × $10^6$ cells was selected for dataset generation as the point of diminishing returns between accuracy and cost. A separate iteration convergence study, also conducted for the generalised model, confirmed that all monitored quantities converged between 200 and 250 iterations; a conservative cutoff of 300 iterations per simulation was adopted. Simulations were executed on the GADI high-performance computing system at the National Computational Infrastructure (NCI), Australia.

\subsection{Dataset Parametrisation and Composition}
\label{sec:DatasetParameters}

The RANS dataset spans a systematic two-parameter space of XLUUV operating conditions: forward speed from 0.10 to 5.00 m/s in increments of 0.10 m/s, and heading angle from 0° (dead ahead) to 50$\deg$ in increments of 5$\deg$. The speed range is intentionally broader than typical LAR operating conditions, extending to 5.00 m/s to capture the full Reynolds number range relevant to XLUUV operations and to ensure the dataset is not artificially constrained to benign, low-speed regimes. At the upper end, the freestream Reynolds number, based on the XLUUV hull length of 22 m, reaches Re $\approx$ 1.09 × 108. The heading angle range captures the progression from a symmetric, dead-ahead wake structure (0$\deg$) through to highly asymmetric wakes with strong cross-flow-induced vortical structures (50$\deg$), which represent the most challenging navigational conditions. The propeller tip speed and disk pressure jump are varied with operating speed as described in Section 3.3.2, so that the propulsive effect of the propeller is correctly scaled to each condition throughout the dataset.

The total original dataset comprises 1,091 RANS simulations covering this parameter space. Data augmentation, and specifically yaw-axis rotation to generate port-side turning cases from starboard cases, and lateral mirroring of dead-ahead (0°) cases was applied to expand the dataset to 4,364 instances \cite{WAKESET}. Of the full dataset, 550 flow conditions, representing 50 distinct speed values paired with 10 distinct angles, sampled to achieve a balanced distribution across the parameter space, were selected for the path planning experiments in this work. This subset was chosen to make the 19,800-trajectory benchmark computationally tractable while maintaining full coverage of the speed and angle parameter space. The distribution of the 550 selected conditions is shown in Table \ref{tab:SimulationMatrix}.

\begin{table}[!ht]
\begin{center}
\small
\begin{tabular}{lll}
\toprule \toprule
\textbf{Speed}     & \textbf{Angle} & \textbf{Trajectories} \\ \toprule \toprule 
$[0.0, 5.0, 0.1] $ & $ [0, 50, 5] $ & 36 per pairing \\ \toprule \toprule 
\end{tabular}
\caption{Composition of the simulations undertaken to generate the analysis. Provided are the lower, upper and increment of each aspect, as well as the number of trajectories generated for each unique configuration. This matrix was repeated for each approach in question (current only, CFD and the generative models).}
\label{tab:SimulationMatrix}
\end{center}
\end{table}

\subsection{Voxel Grid Representation}
\label{sec:GridRepresentation}
For integration into the path planning framework, the RANS velocity magnitude fields are interpolated from the unstructured CFD mesh onto a regular $128^3$ voxel grid using a nearest-neighbour scheme between CFD mesh vertices and grid cell centres. The voxel grid spans the 155 × 155 × 155 m physical domain, giving a voxel edge length of approximately 1.21 m which is sufficient to resolve the coarse-scale spatial structure of the propeller wake and the boundary layer along the XLUUV hull, which are the features most relevant to energy-aware navigation. The XLUUV body geometry is represented as a set of non-traversable voxels within the grid, and the payload bay goal cell is the voxel at the geometric centre of the bay opening.

The 128³ resolution represents a substantial advance over the 48³ resolution of the prior state-of-the-art 3D cGAN surrogate of Yousif et al. \cite{Existing3DSota}, from which the 2D3DGAN architecture used in this work is adapted. At $128^3$, the volume contains 2,097,152 voxels. This is a factor of approximately 19 more than the $48^3$ baseline. This enables the cGAN to capture wake structure features at spatial scales that are operationally relevant to the LAR manoeuvre. The 26-connected voxel adjacency (permitting movement to any face-, edge-, or corner-sharing neighbour) provides the A* planner with the full six-degree-of-freedom spatial freedom needed to route around three-dimensional wake structures. The nearest-neighbour interpolation approach introduces a spatial quantisation error bounded by half the voxel edge length (0.6 m), which is small relative to the characteristic length of the wake structures of interest and does not materially affect the energy cost comparisons reported in Section 7.

The current-only field used as the naive planning baseline is derived from each CFD volume by replacing all per-voxel velocity values with the volume-mean velocity vector, retaining the correct freestream speed and direction while removing all localised wake structure. This provides a clean, unambiguous baseline: any performance difference between a planner operating on this field and one operating on the full CFD field is attributable entirely to the presence and utilisation of wake information.

\section{Path Planning Framework}
\label{sec:Path_Planning_Framework}

The path planning environment is defined as a three-dimensional voxel grid $G \subset \mathbb{Z}^3$ centered on the XLUUV payload bay, spanning the 155 $\times$ 155 $\times$ 155m physical domain described in Section \ref{sec:GridRepresentation}. The grid is discretised into a $128^3$ structure in which each cell corresponds to a planning node $n$ with integer coordinates $n = (x_n, y_n, z_n) \in \{1, ..., 128\}^3$ . Voxels coinciding with the solid XLUUV geometry are designated non-traversable; all remaining nodes constitute the traversable planning graph. Adjacency is defined by the 26-connected neighbourhood, permitting diagonal, edge-crossing, and face-crossing transitions between any pair of nodes that share a face, edge, or corner. The Euclidean distance between node $n$ and a neighbouring node $n'$ is given by Equation \ref{eqn:EuclideanDistance}



\begin{equation}
    \label{eqn:EuclideanDistance}
    d(n, n') = \sqrt{(x_{n'}-x_{n})^2 + (y_{n'}-y_{n})^2 + (z_{n'}-z_{n})^2}
\end{equation}

\subsection{Energy-Weighted Cost Function}

The traversal cost between nodes $n$ and $n'$ is derived from the hydrodynamic drag force the AUV must overcome upon entering the destination node $n'$. This is based on the estimated energy determined via the environmental model - being either current only, RANS CFD, or a cGAN prediction. For a vehicle with drag coefficient $C_D$ and frontal area $A$ navigating through a fluid of density $\rho$, the drag force at node $n'$ is given by Equation \ref{eqn:Force_of_Drag}.


\begin{equation}
    \label{eqn:Force_of_Drag}
    F_D(n') = \frac{1}{2} \rho \left[ v (n') \right]^2 C_D A
\end{equation}

Where $v(n')$ is the velocity magnitude at node $n'$, drawn from the active environmental model active for the planner variant under evaluation: the volume-mean current, the ground truth RANS field, or a cGAN-predicted field (Section \ref{sec:GAN_Pipeline}). The energy cost of the transition from $n$ to $n'$ is the mechanical work done against this drag force over the intervening Euclidean distance, as given by Equation \ref{eqn:Energy_Cost}.

\begin{equation}
    \label{eqn:Energy_Cost}
    E_C(n, n') = F_D(n') \times d(n, n')
\end{equation}

This formulation assigns a higher traversal cost to entering a high-velocity voxel than departing from one, incentivising the planner to route trajectories through velocity-deficit corridors, such as the leeward wake margins and the separated flow region behind the XLUUV hull, even where doing so requires a geometrically longer path. The physical AUV parameters used to evaluate Equation \ref{eqn:Force_of_Drag} are consistent with prior experimental and numerical characterisation of the vehicle class \cite{ZacPlanner}: $C_D = 0.150$ and $A = 0.051m^2$, representative of the GRAALtech X300 platform. Water density $\rho$ is taken as $1025.16 kg\cdot m^{-3}$. 

The cumulative cost of a path $P=[n_0, n_1, ... , n_k]$ from start node $n_0$ to goal node $n_k$ is the g-score, obtained by summing the edge costs along the path in accordance with Equation \ref{eqn:Path_Cost}.

\begin{equation}
    \label{eqn:Path_Cost}
    g(P) = \sum_{i=0}^{n} E_c(n_i, n_{i+1})
\end{equation}

The path planning problem is formulated as the minimisation of g over all valid paths in the traversable graph, as denoted by Equation \ref{eqn:Optimisation_Goal}, where $\mathcal{P}(n_0, n_k)$ is used to denote the set of possible, valid paths from $n_0$ to $n_k$. 

\begin{equation}
    \label{eqn:Optimisation_Goal}
    P^* = \underset{P \in \mathcal{P}(n_0, n_k)}{\operatorname{arg\,min}} \ g(P) = \underset{P = [n_0, ..., n_k]}{\operatorname{arg\,min}} \left( \sum_{i=0}^{k} E_c(n_i, n_{i+1}) \right)
\end{equation}

\subsection{A* Search Algorithm and Heuristic Admissibility}

The A* algorithm solves Equation \ref{eqn:Optimisation_Goal} by maintaining an open set $O$ which is a priority queue of candidate nodes ordered by their total priority score $f(n) = g(n) +h(n)$, and a set of fully expanded nodes. Starting from $n_0$, the algorithm iteratively selects the node with the minimum f-score from $O$, evaluates its 26-connected neighbours $n'$, computes their tentative g-scores and inserts or updates them in $O$ if a cheaper path has been found. The process terminates when the goal node $n_k$ is extracted from $O$, at which point the optimal path is reconstructed via backtracking from $n_k$ through the stored parent pointers. 

A* recovers the globally minimum-cost path if and only if the heuristic $h(n)$ is admissible. That is true when $h(n) \leq h^*(n)$ for all $n$, where $h^*(n)$ denotes the true remaining cost from $n$ to $n_k$ along the optimal path \cite{ZacPlanner}. The heuristic adopted in this work is denoted by Equation \ref{eqn:Heuristic}. 

\begin{equation}
    \label{eqn:Heuristic}
    h(n) = \overline{E_D} \times d(n, n_k)
\end{equation}

In this formulation $d(n, n_k)$ is the Euclidean distance from $n$ to the goal and $\overline{E_D}$ is the global mean energy cost per unit distance, computed as the domain-wide mean of $F_D(n')$ over all traversable nodes in $G$ for the active flow field. The admissibility of this heuristic is established as follows. The true optimal remaining cost $h^*(n)$ is the minimum cumulative drag work over all paths from $n$ to $n_k$. The minimum drag force at any individual node is $\frac{1}{2} \rho (v_{min})^2 C_D A$, where $v_{min}$ is the minimum velocity among all traversable nodes. Since $\overline{E_D}$ is the average over all drag forces, including both the highest and lowest velocity regions, it is necessarily no smaller than $v_{min}^2$-based cost per unit distance. Consequently, $h(n) = \overline{E_D} \times d(n, n_k)$ does not systematically overestimate $h^*(n)$ for any node, which satisfies the admissibility condition. The heuristic provides a computationally inexpensive lower bound on remaining cost that combines geometric proximity to the goal with a domain-representative estimate of the hydrodynamic resistance per unit length, effectively steering A* towards the goal without sacrificing optimality guarantees \cite{ZacPlanner}.

\subsection{Planner Variants}

Four planner variants are evaluated, differing solely in the environmental model from which $v(n')$ is drawn in Equation \ref{eqn:Force_of_Drag}. The first variant, the current-only planner, sets $v(n')$ equal to the volume-mean velocity of the RANS field for the corresponding flow condition, applied uniformly to all traversable nodes. This eliminates all localised wake structure while preserving the correct freestream speed and direction, providing an unambiguous lower-fidelity baseline against which the benefit of hydrodynamic knowledge can be measured. The second variant, the RANS CFD planner, draws $v(n')$ directly from the interpolated ground-truth velocity magnitude field for the given operating condition, representing the upper bound of available environmental information. The third and fourth variants, the PatchGAN GradNorm planner and the 2D3DGAN SA planner, draw $v(n')$ from the $128^3$ voxel velocity fields generated by their respective cGAN pipelines (Section \ref{sec:GAN_Pipeline}), and represent the practically deployable surrogate-informed alternatives to ground-truth CFD.

In addition to the four A* planner variants, two feedforward neural network path approximators from prior work \cite{ZacPlanner}: a current-informed NN (Current\_NN) and a CFD-informed NN (CFD\_NN) are retained as reference comparison points in the results tables. These networks were trained to directly predict AUV waypoint sequences from the scalar inputs $(v, \theta)$ and start/goal coordinates, approximating the trajectory distributions of the current-only and RANS CFD A* planners respectively, without generating or querying an environmental flow field at inference. Their architecture, training procedure, and trajectory-level performance are described in full in \cite{ZacPlanner}; only their evaluation metrics are reproduced here to contextualise the performance of the cGAN-informed planners within the broader landscape of computationally tractable path generation approaches. At inference, both networks produce a complete 130-waypoint trajectory in under 0.4 ms. This is approximately five orders of magnitude faster than any A* variant, but at a cost of trajectory optimality that is quantified alongside the A* results in Section \ref{sec:Results}.

\subsection{Experimental Design and Trajectory Generation}

Trajectories are generated across all 550 distinct flow conditions in the planning dataset, spanning the speed–angle parameter space described in Section \ref{sec:DatasetParameters}. For each condition, 36 AUV start nodes are drawn by uniform random sampling of the rear boundary face of the domain (x = 128 in the grid coordinate system), subject to the constraint that sampled nodes do not coincide with XLUUV body voxels. The same 36 start nodes are used for all four planner variants at each condition, ensuring that inter-planner comparisons reflect differences in planning quality attributable to the environmental model rather than differences in start-point geometry. The goal node $n_k$ is fixed at the voxel at the geometric centre of the payload bay opening across all conditions and planner variants. The resulting 19,800 trajectories per planner (550 conditions $\times$ 36 start nodes) form the complete evaluation set. Across all planners, energy expenditure is computed using ground-truth RANS velocity fields, not the surrogate field used during planning, so that reported energy metrics reflect actual physical energy expenditure in the real flow environment.

\section{GAN Prediction Pipeline}
\label{sec:GAN_Pipeline}

This section describes the hierarchical cGAN pipeline used to generate the predicted flow fields that serve as environmental models for the path planner. The pipeline operates in two stages: a 2D cGAN that synthesises the conditional planar slice inputs from sparse scalar metadata, followed by a 3D cGAN that lifts these slices to a full volumetric field. The two 3D architectures selected for benchmarking — the Regularised PatchGAN GradNorm and the 2D3DGAN with Self-Attention — are described at summary level, and their inference characteristics relative to ground-truth CFD are reported.

\subsection{Motivation}

The fundamental challenge motivating this pipeline is that the 3D cGAN generators require, as conditional input, two orthogonal 2D planar slices through the flow field — one horizontal and one vertical — together with the scalar operating condition metadata (vehicle speed v and heading angle $\theta$). In a deployed scenario, these planar slices would not be directly available from onboard sensors. The hierarchical pipeline resolves this by using a pre-trained 2D cGAN (Stage 1) to synthesise both required slices from the scalar metadata alone, so that the full chain from {$v$, $\theta$} → 2D slices → 3D volume can be executed onboard without any requirement for external flow measurements. The complete pipeline is illustrated in Figure \ref{fig:GAN_Planner_Pipeline} and proceeds as follows:

\begin{figure}[!ht]
\centering
\includegraphics[width=0.8\linewidth]{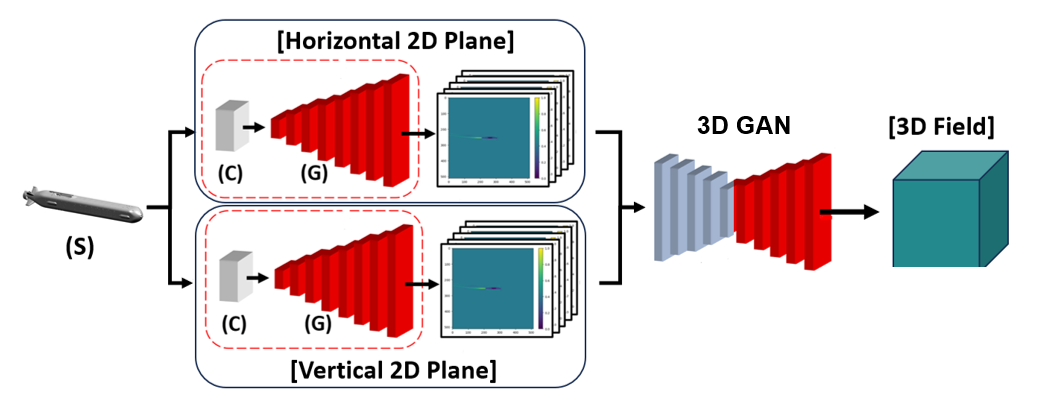} 
\caption{Conceptual pipeline for path planning in a predicted flow field. Sparse speed and angle data is used (S) in the 2D Model 4 network \cite{Zac_PhD_Thesis} to produce a vertical and horizontal 2D planar slice, using the pretrained model. These 2D planar slices and metadata (S) are fed into a pre-trained 3D GAN generator (e.g., PatchGAN or 2D3DGAN SA). The generator outputs a full $128^3$ volumetric flow field, for a given speed and angle (S), which then serves as the environment model for the A* path planner to compute a trajectory.}
\label{fig:GAN_Planner_Pipeline}
\end{figure}

\begin{itemize}
    \item \textbf{Stage 1 (2D slice synthesis)}: The scalar pair ($v$, $\theta$) is passed through a pre-trained 2D conditional deep convolutional GAN (cDCGAN), specifically the Model 4 architecture identified as optimal in prior work \cite{Zac_PhD_Thesis}. This network generates the horizontal and vertical $128 \times 128$ pixel velocity magnitude slices that characterise the flow field at the given operating condition. The 2D generator produces these slices at a mean inference time of approximately 16 $\mu s$ \cite{Zac_PhD_Thesis}, adding negligible latency to the overall pipeline.

    \item \textbf{Stage 2 (3D volume generation)}: The two synthesised 128 × 128 planar slices and the scalar metadata {$v$, $\theta$} are passed through one of the pre-trained 3D cGAN generators (PatchGAN GradNorm or 2D3DGAN SA, described in Sections 4.2 and 4.3). The generator produces a $128^3$ voxel velocity magnitude volume representing the full 3D flow field. This volume is then interpolated onto the planning grid and used directly as the environmental model by the A* path planner (Section \ref{sec:Path_Planning_Framework}).
\end{itemize}

This two-stage design decouples the complexity of 3D volume generation from the sparsity of onboard sensing. The scalar inputs ($v$, $\theta$) can be estimated in real time from standard AUV navigation sensors — inertial measurement units, Doppler velocity logs, and acoustic positioning — without any direct flow measurement. The pipeline therefore provides a practically deployable route to wake-informed path planning that is not contingent on sparse flow sensing infrastructure. The complete end-to-end inference time, from scalar inputs to a populated $128^3$ planning grid, is dominated by the 3D generator stage and ranges from approximately 12 to 130 $\mu s$ depending on the architecture selected (Section \ref{sec:GAN_Pipeline}), compared to RANS simulation times on the order of hours on high-performance computing hardware.

\subsection{Slice Synthesis}

The 2D cGAN used for planar slice synthesis is the Model 4 hybrid architecture described in \cite{Zac_PhD_Thesis}, which combines a Self-Attention GAN (SAGAN) generator \cite{zhang2019self} with a PatchGAN discriminator \cite{isola2017image}. The generator receives a concatenation of a 64-dimensional random latent vector z and a two-element metadata embedding of ($v$, $\theta$). Self-attention layers within both the generator and discriminator enable the network to relate spatially distant features of the velocity field. This is critical for accurately reconstructing the far-field wake structure and its relationship to the propeller region without being limited by the local receptive field of a purely convolutional architecture. The PatchGAN discriminator evaluates overlapping local patches of the generated image rather than a single global score, providing a spatially dense adversarial training signal that enforces local structural fidelity and suppresses the blurring artefacts characteristic of global discriminators.

The network is trained on 512 × 512 velocity magnitude images drawn from the CFD dataset (Section 3.4), with bilinear downsampling to the 128 × 128 conditioning resolution prior to use in the 3D pipeline. Training uses a composite loss combining adversarial hinge loss, MSE and the custom Gradient-Weighted MSE (GMSE) loss \cite{ZacLoss}. This loss applies locally elevated penalty weights at high-gradient regions including the propeller wake boundary and the hull boundary layer. This is then further supported by structural similarity (SSIM) and perceptual losses. In the optimised configuration, the model achieves a PSNR of 63.2 dB, SSIM of 0.999, and FID of 25.8 on the test set \cite{Zac_PhD_Thesis}, with a mean inference time of 16.1 $\mu s$ and a memory footprint of approximately 1.2 GB in mixed precision, making it deployable on modern embedded GPU hardware.

\subsection{3D Volume Generation}

\begin{figure}[H]
    \centering
    \subfloat[]{\includegraphics[height=1in]{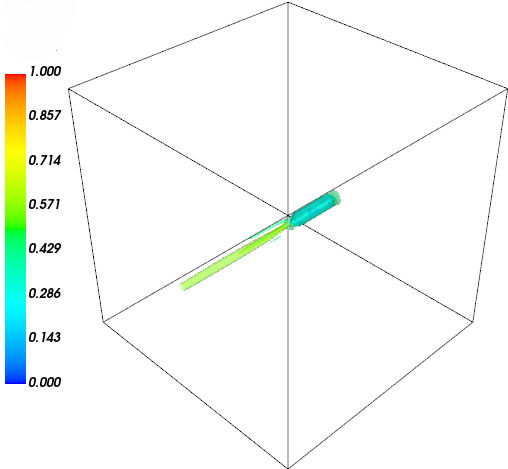}%
    \label{fig:3D_cases_real_data_A}}
    \hfil
    \subfloat[]{\includegraphics[height=1in]{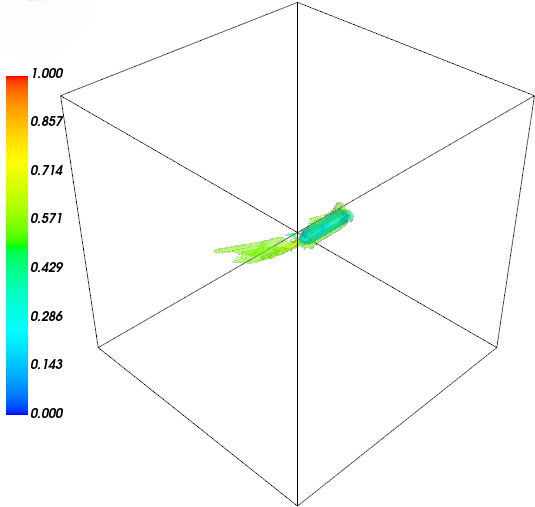}%
    \label{fig:3D_cases_real_data_B}}
    \hfil
    \subfloat[]{\includegraphics[height=1in]{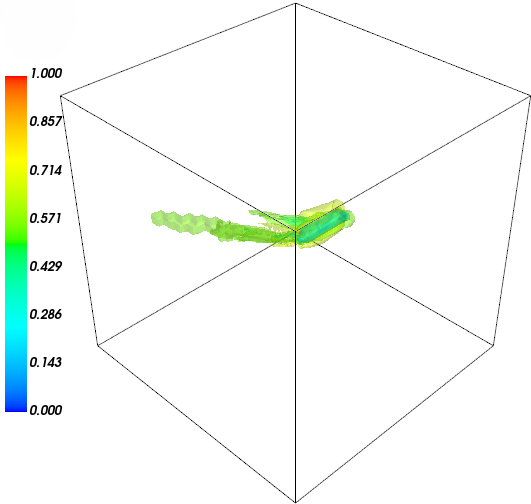}%
    \label{fig:3D_cases_real_data_C}} 
    \hfil
    \subfloat[]{\includegraphics[height=1in]{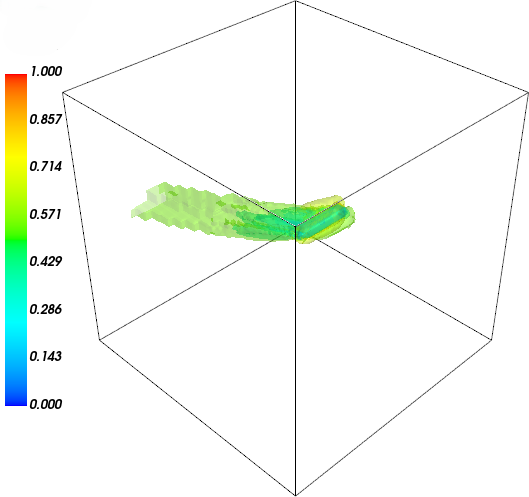}%
    \label{fig:3D_cases_real_data_D}} 
    \\

    \subfloat[]{\includegraphics[height=1in]{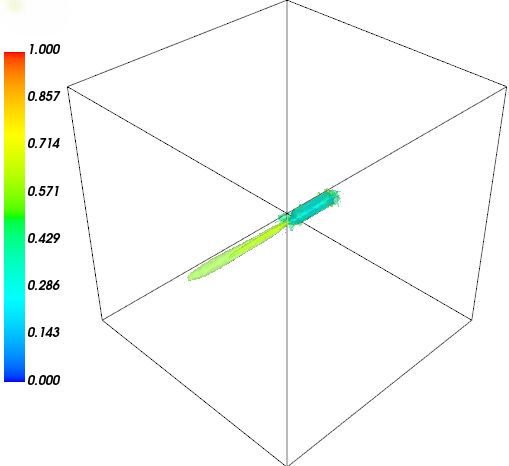}%
    \label{fig:3D_cases_PatchGAN_data_A}}
    \hfil
    \subfloat[]{\includegraphics[height=1in]{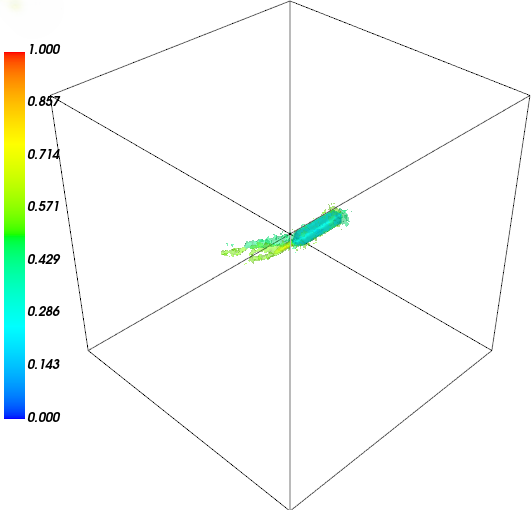}%
    \label{fig:3D_cases_PatchGAN_data_B}}
    \hfil
    \subfloat[]{\includegraphics[height=1in]{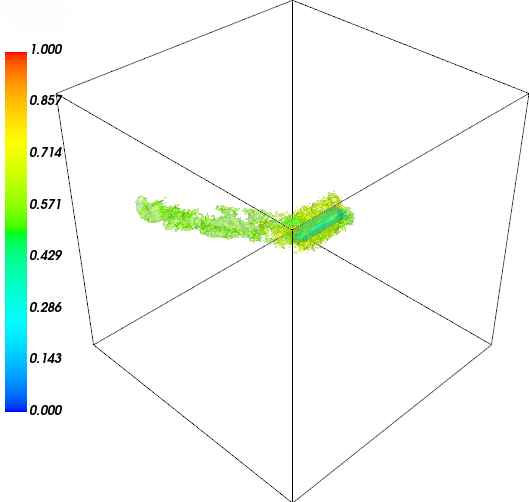}%
    \label{fig:3D_cases_PatchGAN_data_C}}
    \hfil
    \subfloat[]{\includegraphics[height=1in]{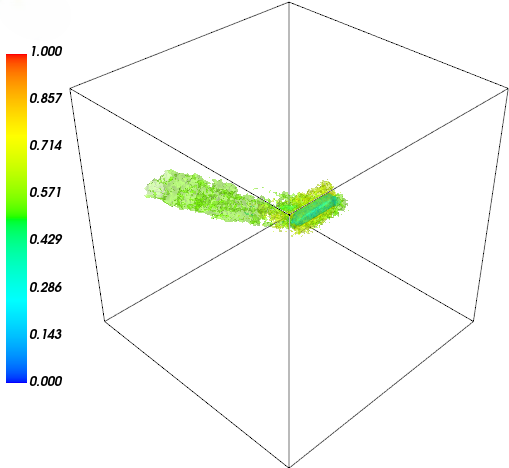}%
    \label{fig:3D_cases_PatchGAN_data_D}}
    
    \caption{\textbf{(a - d):} The ground truth flow fields from RANS CFD results for a speed of 2.5 m/s with turns of 0, 10, 20 and 40 degrees respectively. \textbf{(e - h):} show the generated fields from the PatchGAN GradNorm model. All velocity magnitude data has been normalised between to the range [0,1].}
    \label{fig:3D_volumes_data}
\end{figure}

Two 3D cGAN architectures are used as alternative Stage 2 generators in this work. They represent contrasting points on the fidelity–efficiency trade-off space and were selected from a broader development programme \cite{Zac_PhD_Thesis} on the basis of their overall composite performance score across reconstruction accuracy, perceptual quality, and computational cost. Both generators share the same conditional input structure — two 128 × 128 planar slices (horizontal and vertical), the scalar metadata ($v$, $\theta$), and a 128-dimensional latent vector — and produce a $128^3$ voxel velocity magnitude volume as output. Both are trained with the composite loss function incorporating adversarial hinge loss, MSE, SSIM, GMSE, and a perceptual slice loss computed via a pre-trained VGG-19 network applied to cross-sectional slices of the generated volume \cite{Zac_PhD_Thesis}. Discriminator training uses an R1 gradient penalty \cite{Mescheder2018} combined with gradient clipping to stabilise training against the well-documented failure modes of 3D adversarial training at high resolution. Shown in Figure \ref{fig:3D_volumes_data} are the ground truth RANS CFD fields for selected instances, with high degrees of separation, as well as PatchGAN predicted fields for comparison.

\subsubsection{Regularised PatchGAN GradNorm}

The Regularised PatchGAN GradNorm is the lighter and faster of the two 3D generators, and is the primary practical candidate for onboard deployment. Its generator architecture uses two dedicated 2D CNN slice encoders that process the horizontal and vertical input slices independently, each producing a 512-dimensional feature vector. These slice features, together with a metadata embedding of ($v$, $\theta$) processed through a fully connected embedding layer, and the latent vector, are concatenated and projected onto an initial 4 × 4 × 4 3D feature volume via a linear layer. The 3D volume is then progressively upsampled through five transposed convolutional blocks from $4^3$ to $128^3$, using 3 × 3 × 3 kernels to mitigate checkerboard artefacts \cite{TransposedPathway}. Feature-wise Linear Modulation (FiLM) layers \cite{FiLMLayers} are applied in the first three upsampling blocks, modulating the 3D feature maps multiplicatively and additively using the metadata embedding; this provides strong, persistent conditioning throughout the generation process. A 3D cross-attention module, in which queries are derived from the 3D feature maps and keys and values from the combined slice features, is applied at the $32^3$ resolution stage, allowing any spatial location in the evolving volume to attend to any token in the 2D slice feature sequences. This cross-attention is computationally more tractable than full 3D self-attention — scaling linearly in the 3D feature map dimension rather than quadratically — while still capturing long-range spatial correspondences between the planar slice observations and the 3D volume.

The discriminator is a 3D PatchGAN \cite{isola2017image} that outputs a 4 × 4 × 4 grid of local realism scores rather than a single global classification. It uses spectral normalisation \cite{miyato2018spectral} throughout its convolutional blocks and incorporates a 3D global self-attention layer, providing both local discriminative feedback and global structural coherence assessment. Gradient normalisation of the discriminator loss (hence the GradNorm designation) is applied via the R1 penalty and gradient clipping, stabilising training against discriminator collapse.

In the optimised configuration (full training, 550 epochs, learning rate $4 \times 10^{-5}$, batch size 4), the Regularised PatchGAN GradNorm achieves MSE = $1.22 \times 10^{-4}$, PSNR = 43.6 dB, SSIM = 0.941, FID = 51.9, and LPIPS = 0.039. Its composite normalised score of 0.916 is the highest among all architectures evaluated when inference time and memory are included in the score \cite{Zac_PhD_Thesis}. The generator inference time is 12.0 $\mu s$ (mixed precision FP16/FP32) with a memory footprint of 1.21 GB. Qualitative evaluation demonstrates robust volumetric reconstruction across the full speed and angle parameter space, including accurate reproduction of the spatially separated wake structure at high heading angles and the propeller race at low angles, with average field divergence below 10\% of the maximum recorded velocity in most operating conditions \cite{Zac_PhD_Thesis}.

\subsubsection{2D3DGAN with Self-Attention}

The 2D3DGAN SA is an adaptation of the architecture proposed by Yousif et al. \cite{Existing3DSota} which demonstrated 3D flow reconstruction at $48^3$ resolution. This has been extended and modified to operate at $128^3$ \cite{Zac_PhD_Thesis}. It represents the upper bound of volumetric reconstruction fidelity available within this study, at the cost of substantially higher computational requirements. The generator uses two 2D CNN encoders to process the input slices into feature maps at 16 × 16 spatial resolution. The 256-token flattened representations of both slices are fused via cross-attention, allowing locations in each slice to directly attend to locations in the other — capturing the 3D spatial coherence that must be inferred from the two orthogonal observations. The fused 2D features are then reshaped and lifted to an initial 3D volume before progressive upsampling to $128^3$ through transposed convolutional blocks.

The defining architectural feature is a global 3D self-attention layer placed mid-path through the generator at the $32^3$ feature resolution (64 channels). This attention operates over all (32 × 32 × 32) = 32,768 spatial tokens simultaneously, enabling any voxel in the intermediate feature volume to attend to any other. The resulting global receptive field allows the generator to model long-range hydrodynamic relationships such as the propagation of propeller-induced swirl into the far-field wake. Local convolutional operations cannot capture without very deep networks. A matching self-attention layer is incorporated into the discriminator at the same feature resolution. The computational cost of this mechanism scales quadratically with the spatial dimension of the feature map, O((DHW)2), which at $32^3$ yields an attention matrix requiring approximately 25.7 GB of GPU memory and an inference time of 130 $\mu$s - both substantially higher than the PatchGAN GradNorm.

In the optimised configuration (learning rate $4 \times 10^{-5}$, batch size 2, $\lambda_{MSE}$ = 200, $\lambda_{GMSE}$ = 200), the 2D3DGAN SA achieves MSE = $8.38 \times 10^{-6}$, PSNR = 53.2 dB, SSIM = 0.992, FID = 12.2, and LPIPS = 0.023. These reconstruction metrics represent the highest fidelity achieved by any architecture in this study, particularly in terms of SSIM and FID, which reflect structural coherence and distributional similarity to the ground truth CFD fields \cite{Zac_PhD_Thesis}. However, qualitative volumetric comparisons \cite{Zac_PhD_Thesis} reveal that the 2D3DGAN SA shows localised under-prediction of the propeller wake velocity deficit at low operating speeds and low heading angles - conditions where the wake structure is subtle but energetically significant - and can smooth vortical structures at moderate cross-flow angles. The PatchGAN GradNorm, despite lower pixel-level reconstruction scores, demonstrates more spatially consistent wake structure reproduction across the full parameter space. This distinction motivates the inclusion of both architectures in the path planning benchmark: the 2D3DGAN SA provides an upper bound on reconstruction-metric performance while potentially lagging in planning-relevant spatial features, and the PatchGAN GradNorm provides an operationally viable surrogate with more robust structural fidelity.

\subsection{Inference Characteristics and Deployment Feasibility}

Table \ref{tab:InferenceTiming} summarises the inference timing and memory characteristics of all components of the prediction pipeline, measured on an NVIDIA A6000 48~GB GPU in mixed precision (FP16/FP32). All timings represent the mean over 1,000 inference calls with GPU cache warm-up. The table is structured in three tiers: individual stage inference times for the two cGAN stages in isolation, A* path planning search times for each environmental model (evaluated separately, after field generation), and the combined end-to-end pipeline times. For reference, the approximate wall-clock time for a single RANS CFD simulation is also provided, representing the ground-truth data generation cost the pipeline is designed to replace.

\begin{table}[!ht]
\centering
\small
\setlength{\tabcolsep}{5pt}
\begin{tabular}{llrrl}
\toprule \toprule
\textbf{Component} & \textbf{Path} & \textbf{Time} & \textbf{Memory} & \textbf{Source} \\
\midrule
\multicolumn{5}{l}{\textit{Stage 1 — 2D slice synthesis (shared by both paths)}} \\
\midrule
2D cDCGAN (Model~4) & Both & 16.1~$\mu$s & 1.21~GB & \cite{Zac_PhD_Thesis} \\
\midrule
\multicolumn{5}{l}{\textit{Stage 2 — 3D volume generation}} \\
\midrule
PatchGAN GradNorm generator & PatchGAN & 12.0~$\mu$s & 1.21~GB & \cite{Zac_PhD_Thesis} \\
2D3DGAN SA generator        & 2D3DGAN~SA & 130.4~$\mu$s & 25.68~GB & \cite{Zac_PhD_Thesis} \\
\midrule
\multicolumn{5}{l}{\textit{End-to-end cGAN inference (Stage 1 + Stage 2, field only)}} \\
\midrule
Combined pipeline & PatchGAN   & \textbf{28.1~$\mu$s} & 2.42~GB & — \\
Combined pipeline & 2D3DGAN~SA & \textbf{146.5~$\mu$s} & 26.89~GB & — \\
\midrule
\multicolumn{5}{l}{\textit{A* path planning search (mean across all 550 flow conditions)}} \\
\midrule
A* search & Current-only & 68.2~s & — & \S\ref{sec:Results} \\
A* search & CFD          & 55.7~s & — & \S\ref{sec:Results} \\
A* search & PatchGAN     & 44.9~s & — & \S\ref{sec:Results} \\
A* search & 2D3DGAN~SA   & 45.4~s & — & \S\ref{sec:Results} \\
\midrule
\multicolumn{5}{l}{\textit{Full planning pipeline (cGAN inference + A* search)}} \\
\midrule
cGAN + A* & PatchGAN   & 28.1~$\mu$s $+$ 44.9~s & 2.42~GB & — \\
cGAN + A* & 2D3DGAN~SA & 146.5~$\mu$s $+$ 45.4~s & 26.89~GB & — \\
\midrule
\multicolumn{5}{l}{\textit{Reference — ground-truth CFD (per operating condition)}} \\
\midrule
RANS CFD simulation & — & $\mathcal{O}$(hours) on HPC & N/A & \cite{SCA24} \\
\bottomrule \bottomrule
\end{tabular}
\caption{Inference timing and memory footprint for all components of the cGAN prediction pipeline. Stage~1 (2D cDCGAN) and Stage~2 (3D generator) timings are measured in isolation on an NVIDIA A6000 48~GB GPU, mixed precision FP16/FP32, mean over 1,000 warm-cache calls. A* search times are means across all 550 flow conditions and 36 start points (19,800 trajectories per planner), drawn from Table~\ref{tab:Compute_Time}. The full pipeline time is dominated by A* search; cGAN inference contributes less than 0.1\% of total planning time in both paths. RANS CFD simulation time is condition-dependent and cluster-dependent; values reported here are representative of the GADI NCI computing environment used in this work~\cite{Zac_PhD_Thesis}.}
\label{tab:InferenceTiming}
\end{table}

Both cGAN pipeline paths generate a complete $128^3$ planning field in under 150~$\mu$s which is more than four orders of magnitude faster than a single RANS CFD simulation. Crucially, however, Table~\ref{tab:InferenceTiming} reveals that cGAN inference is not the computational bottleneck of the planning pipeline: it contributes less than 0.1\% of total planning time. The dominant cost is the A* search itself, which requires 44.9~s (PatchGAN path) and 45.4~s (2D3DGAN~SA path) on average. This has two important implications. First, the deployment feasibility question is not primarily a question of cGAN inference speed, as both architectures are fast enough, but of the memory required to load and run the 3D generator on available hardware. Second, the A* search time provides an indirect measure of cost-surface structure: a more navigable cost landscape, with smoother gradients, enables A* to converge faster, which is why both GAN planners outperform the CFD planner in mean search time (44.9--45.4~s vs 55.7~s) despite generating less accurate fields.

The PatchGAN GradNorm path, with a combined cGAN memory footprint of 2.42~GB and an end-to-end field generation time of 28.1~$\mu$s, is readily deployable on hardware such as modern embedded GPU accelerators like the NVIDIA Jetson AGX series (~16--32~GB unified memory). Such systems are increasingly used in advanced AUV computing payloads. The 2D3DGAN~SA path requires 25.68~GB for the 3D generator alone, exceeding the capacity of current embedded GPU hardware. Its practical deployment is therefore limited to pre-mission or surface-assisted planning scenarios, where a dedicated high-performance GPU is available, or to offline benchmarking contexts. Both paths achieve field generation rates that impose no latency constraint on the A* planning cycle; the sub-millisecond inference time means that the predicted field can be regenerated for any new operating condition in real time, with the A* search remaining the sole computational bottleneck.

Comparing the GAN planning pipelines against the CFD reference underscores the deployment asymmetry that motivates this work. A CFD-informed A* planner requires a pre-computed RANS simulation for each operating condition. This is an $\mathcal{O}$(hours) computation on a high-performance computing cluster that cannot be executed onboard. It is therefore constrained to a pre-computed lookup strategy: the full 550-condition dataset used in this benchmark required 1,091 original RANS simulations~\cite{Zac_PhD_Thesis}, with augmentation to 4,364 instances. The GAN planners, by contrast, generate the required field from scalar metadata in under 150~$\mu$s at any operating condition within the training distribution, including conditions not explicitly in the pre-computed dataset, without any additional computational cost. This generalisation capability, in which a plausible field is produced for any $\{v, \theta\}$ pair, not just the 550 conditions that were simulated is a qualitatively different capability from a lookup table, and is the property that makes onboard, reactive wake-aware planning tractable.

\section{Evaluation Metrics}
\label{sec:Evaluation_Metrics}
The energy capacity and hydrodynamic stability of AUVs is extremely limited. Small scale fluctuations in encountered velocity and turbulence have the potential to offset, change and even significantly alter the trajectory and ability to follow a trajectory of the AUV. To ensure that the generated paths and hydrodynamic information are sufficient, several metrics designed to assess this have been implemented. These metrics assess the trajectories traversal of high velocity regions, turbulent regions, trajectory energy cost and the overall length of the path. The goal of these combined metrics is to determine if the trajectories produced by one set of hydrodynamic information are any better than the other. To allow for this assessment, the path ($P$) is decomposed into a series of sequential points, as denoted by Equation \ref{eqn:PathDecomp}.

\begin{equation}
    \label{eqn:PathDecomp}
   P = \{ n_1, n_2, n_3, \dots, n_N \} 
\end{equation}

\subsection{Path High Velocity Regions}
\label{sec:HighVelocityRegions}

To make an assessment of whether a high velocity cell is encountered, a velocity threshold ($u_{threshold}$) is used. This threshold assesses if the velocity of a given cell ($n$) on the path is higher than one standard deviation ($\sigma_u$) from the median velocity ($\overline{u}$). This threshold is denoted by Equation \ref{eqn:v_threshold}.

\begin{equation}
    \label{eqn:v_threshold}
    u_{threshold} = \overline{u} + \sigma_u
\end{equation}

To make an assessment of the path quality, the number of these high velocity cells is summed in accordance with Equation \ref{eqn:PathHighVelocity} and \ref{eqn:PathVelocityCutOff} where Equation \ref{eqn:PathHighVelocity} denotes the sum ($N$) of the high velocity cells encountered on a given path $P$. $\delta_{\text{V}}$ denotes the threshold at which a given cell on the path ($n_i$) is deemed to be counted, and thus summed within Equation \ref{eqn:PathHighVelocity}.

\begin{equation}
\label{eqn:PathHighVelocity}
    N_{\text{high-velocity}}(P) = \sum_{i=1}^{N} \delta_{\text{V}}(n_i)  
\end{equation}

\begin{equation}
\label{eqn:PathVelocityCutOff}
    \delta_{\text{V}}(n_i) =
    \begin{cases}
    1, & \text{if } v(n_i) \geq u_{\text{threshold}} \\
    0, & \text{otherwise}
    \end{cases}
\end{equation}

\subsection{Path Turbulent Regions}
\label{sec:TurbulentRegions}

In an ideal case, the majority of the AUV trajectory would be executed in a consistent flow field, free of velocity fluctuation. In such an ideal case, the control effectiveness of the vehicle is retained, relatively constant, and control actions exert an expected change in pitch, yaw and roll. However, in the real world, and specifically in applications involving more than one vehicle, this isn't the case. Small scale flow features, vortices, shed hydrodynamics and propeller race can all cause velocity fluctuations and turbulent changes to the flow. These flow changes and their localised velocity fluctuations impact control effectiveness, hydrodynamics and effective GNC. To make a quantitative assessment of this over a given trajectory, the number of cells where the velocity changes is summed. As the ground truth RANS CFD data may have small scale fluctuations due to numerical precision, a small threshold is applied to the velocity ($\varepsilon$ 0.005, 0.5\% of $u_{freestream}$) that ensures only significant fluctuations are summed. The summation is provided in Equation \ref{eqn:PathTurbulence} and the threshold limits are provided in Equation \ref{eqn:PathTurbulenceCutOff}. 

\begin{equation}
\label{eqn:PathTurbulence}
   N_{\text{turbulent}}(P) = \sum_{i=1}^{N} \delta_{\text{T}}(n_i) 
\end{equation}

The turbulent cell count complements the high-velocity metric by capturing the \textit{character} of the flow environment rather than its peak intensity. Where the high-velocity threshold (Equation~\ref{eqn:v_threshold}) identifies cells in which the mean velocity magnitude is large enough to impose a sustained drag penalty, the turbulent count identifies cells in which the velocity changes appreciably between successive nodes on the path, regardless of the absolute magnitude. Such transitions are operationally significant because they represent localised perturbations to the hydrodynamic load on the vehicle: an abrupt change in oncoming velocity drives a corresponding change in effective angle of attack and incident drag force, exciting pitch, yaw, and roll moments that the vehicle's control system must reject to maintain the planned trajectory. In the propeller wake and shear layers surrounding the XLUUV hull, these transitions are frequent and spatially compact, creating a challenging environment for guidance, navigation and control (GNC) systems even when no individual cell velocity exceeds the high-velocity threshold.

\begin{equation}
\label{eqn:PathTurbulenceCutOff}
    \delta_{\text{T}}(n_i) =
    \begin{cases}
    1, & \text{if } v'(n_i) \geq \varepsilon \\
    0, & \text{otherwise}
    \end{cases}
\end{equation}

The threshold $\varepsilon$ is applied to exclude numerical noise inherent in the discretisation of the RANS solution onto the voxel grid: transitions smaller than 0.5\% of the freestream speed are classified as steady and do not contribute to the count. All transitions at or above this value are counted equally, without further magnitude weighting. This flat counting convention means the metric is sensitive to the \textit{spatial frequency} of velocity transitions along the path --- a property that directly reflects the degree of flow unsteadiness experienced by the vehicle --- rather than to the amplitude of individual events, which is captured by the high-velocity metric. Together, the two metrics provide complementary views of the safety environment: the high-velocity count identifies exposure to the energetically extreme wake core, while the turbulent count characterises exposure to the distributed fluctuation environment of the surrounding shear layers.

\subsection{Path Energy Cost}
\label{sec:PathEnergyCost}

The path energy cost $E(P)$ is the primary optimisation objective of the A* planner and the principal metric by which planner performance is evaluated. It measures the total mechanical work done against hydrodynamic drag along a trajectory, integrating the drag force (Equation~\ref{eqn:Force_of_Drag}) over the Euclidean distance (Equation~\ref{eqn:EuclideanDistance}) of each successive node transition. The drag force at each transition is a function of the change in absolute velocity magnitude between nodes $n_i$ and $n_{i+1}$, the AUV drag coefficient $C_D$, and the vehicle frontal area $A$, making the cost of each step directly sensitive to the local gradient of the flow field. A transition that moves the vehicle from a low-velocity region to a high-velocity region incurs a large drag penalty; a transition in the opposite direction incurs a smaller one.

This formulation captures the core physics of energy-optimal routing in a non-uniform flow: the optimal path is not necessarily the shortest, but the one that minimises cumulative work by exploiting low-velocity corridors and avoiding high-velocity regions. It is because of this velocity-weighting of the cost function that the high-fidelity wake representation provided by the RANS CFD or cGAN-predicted fields is operationally valuable. A planner operating on a uniform current field applies the same velocity at every node in the domain, making the cost of each step proportional to distance alone; it therefore defaults to the shortest path. A planner with access to the spatially resolved wake field can instead identify routes that traverse low-velocity wake corridors or approach the payload bay from angles that minimise head-on exposure to the propeller race, even when those routes are geometrically longer.

\begin{equation}
\label{eqn:PathEnergy}
    E(P) = \sum_{i=1}^{N-1} F_D(n_i, n_{i+1}) \times d(n_i, n_{i+1})    
\end{equation}

The energy cost $E(P)$ is evaluated identically for all planner variants using the ground-truth RANS velocity field, regardless of the field that was used during planning. This ensures that the metric reflects the actual physical energy expenditure the AUV would experience in the real flow environment, not the energy predicted by the surrogate model. Differences in $E(P)$ between planners therefore reflect differences in the quality of the trajectories they generate, not differences in how each planner estimates the cost of its own trajectories.

\subsection{Path Length}
\label{sec:PathLength}

The path length $L(P)$ is the total Euclidean arc length of the trajectory, summed over all node transitions. Unlike the energy cost, it is independent of the flow field and measures only the geometric extent of the path through the voxel grid. It serves as a secondary metric that characterises the geometric trade-off accepted by each planner in pursuit of energy efficiency: a planner that routes the vehicle around energetically costly wake regions will inevitably produce a longer path than one that takes the direct approach, and the magnitude of this extension reflects the scale of the avoidance manoeuvre the planner has chosen to execute.

Reporting $L(P)$ alongside $E(P)$ is necessary because the two metrics can move in opposite directions, and neither alone is sufficient to characterise the quality of a trajectory. A path that is short but traverses the high-velocity wake core may be geometrically efficient but energetically and operationally costly; a path that is long but entirely avoids the wake may be energetically optimal but operationally impractical if mission time constraints are binding. The joint behaviour of $E(P)$ and $L(P)$ across planners reveals the nature of the energy--geometry trade-off that each environmental model induces: the ratio of the percentage energy saving to the percentage path length increase provides a dimensionless measure of routing efficiency that is directly comparable across conditions and planner variants.

\begin{equation}
\label{eqn:PathLength}
   L(P) = \sum_{i=1}^{N-1} d(n_i, n_{i+1}) 
\end{equation}

As with the energy cost, $L(P)$ is evaluated from the planned trajectory coordinates and requires no knowledge of the flow field. It is therefore immune to any inaccuracy in the surrogate model and provides an unambiguous geometric comparison between planner outputs. The path length also provides an indirect diagnostic of wake spatial extent: a planner that routes the vehicle significantly further than the direct path has inferred --- correctly or incorrectly --- that the flow obstacle it must avoid is spatially extensive. Comparing the path length extensions of the GAN and CFD planners therefore reveals whether the surrogate's representation of the wake boundary is consistent with the ground truth in spatial scale, complementing the energy and high-velocity metrics that characterise consistency in velocity magnitude.

\section{Results}
\label{sec:Results}

This section presents a systematic comparative analysis of all six path planning approaches across the five evaluation metrics defined in Section 6. Results are structured across twenty speed–angle bins spanning the full 550-condition parameter space: five speed ranges (0.3–1.0, 1.0–2.0, 2.0–3.0, 3.0–4.0, and 4.0–5.0 m/s) crossed with four heading angle ranges ($0 - 15^o$, $15^o - 30^o$, $30^o - 45^o$ and $45^o - 50^o$). For each bin, reported values are means over all contributing conditions and their 36 start-node trajectories; standard deviations reflect variability across both axes of variation. Complete numerical results are provided in Tables \ref{tab:Energy_Expenditure}–\ref{tab:Compute_Time}. The three-dimensional isosurface plots in Figures \ref{fig:MeanPathEnergy}–\ref{fig:MeanHighVelocityCells} provide continuous visualisations of energy expenditure, path length, and high-velocity cell encounters as functions of the two-dimensional speed–angle space for all four A* planner variants.

The analysis addresses three questions in sequence. Section \ref{sec:HydroKnowledge} characterises the value of perfect RANS CFD wake knowledge relative to the current-only baseline, establishing the upper bound of available planning benefit. Section \ref{sec:SurrogatePerf} quantifies the fraction of that benefit recovered by each cGAN surrogate planner. Section \ref{sec:ArchComp} examines the performance differences between the two surrogate architectures and situates the NN approximators from prior work \cite{ZacPlanner} within this comparison. Sections \ref{sec:CellTraversal} and \ref{sec:CompPerf} address the turbulence trade-off and computational performance characteristics that complete the five-metric evaluation.

\subsection{The Value of Complete Hydrodynamic Wake Knowledge}
\label{sec:HydroKnowledge}

\begin{figure}[H]
    \centering
    \includegraphics[width=1.0\linewidth]{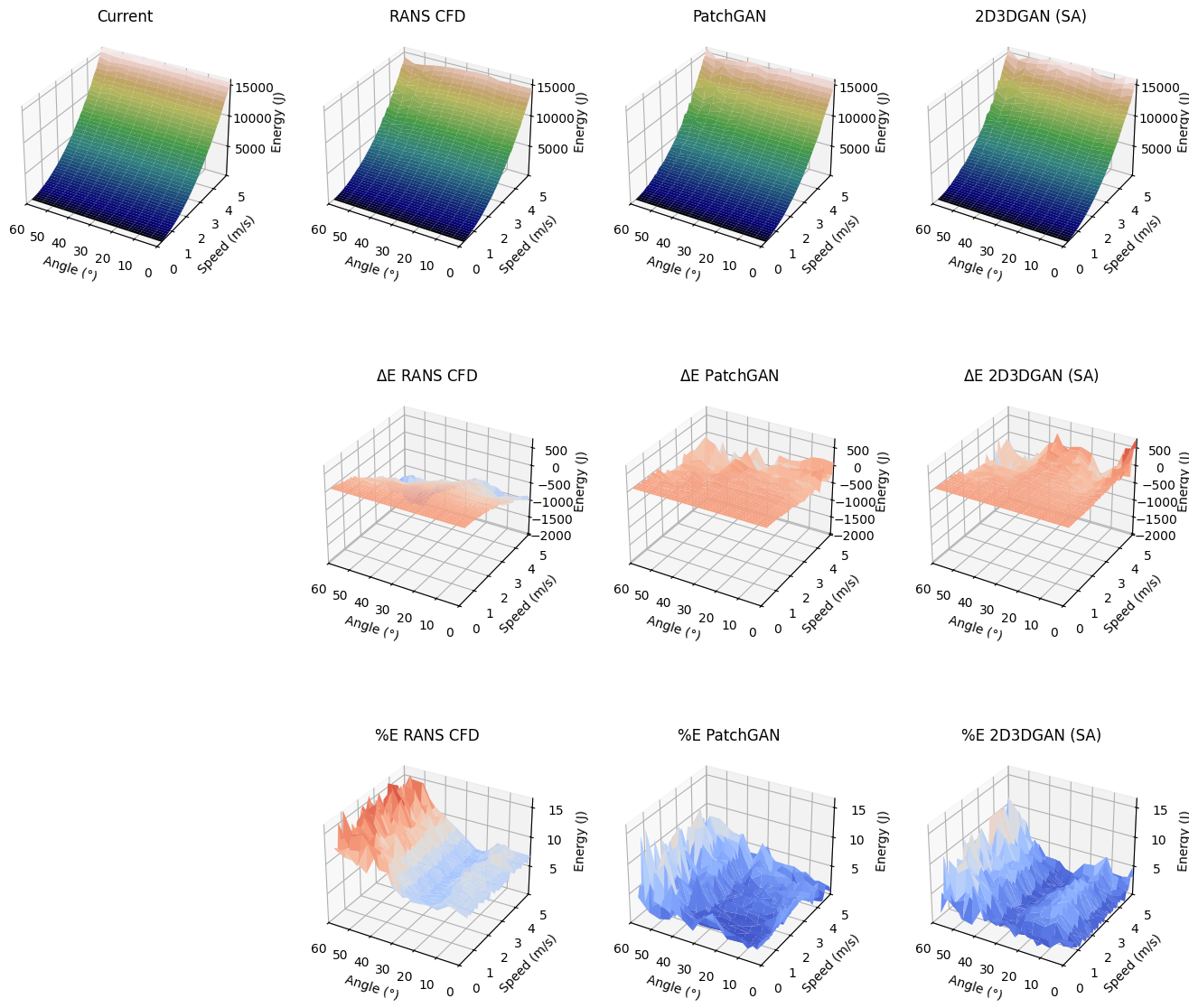}
    \caption{3D isosurface plot of the mean path energy cost (E) for the Current, CFD, PatchGAN and 2D3DGAN (SA) A* planners. Denoted is the angle (x-axis) and the speed (y-axis) of the flow condition that the trajectory was planned on. Vertical (z-axis) used to denote the path energy in joules (row 1), change in path energy (row 2) and the percentage difference (row 3). Differences are given relative to the Current only A* planner.}
    \label{fig:MeanPathEnergy}
\end{figure}

\textit{Energy expenditure}. The RANS CFD planner achieves lower energy expenditure than the current-only baseline across all twenty speed–angle bins, with savings ranging from 5.74\% at 1.0–2.0 m/s, $15^o - 30^o$ to 12.51\% at 0.3–1.0 m/s, $45^o - 50^o$ (Table \ref{tab:Energy_Expenditure}). The dominant pattern is one of strong angular dependence and comparatively weak speed dependence. At near-dead-ahead conditions ($0 - 15^o$), energy savings stabilise in the range of 7.1–7.3\% across all five speed bins, indicating that the navigational benefit of complete wake knowledge is primarily determined by wake asymmetry rather than Reynolds number over the tested range. As heading angle increases to $45^o - 50^o$, savings rise consistently to 11.75–12.51\%, reflecting the progressively greater spatial extent and energetic cost of the asymmetric, separated wake generated by the turning XLUUV. These percentage savings are preserved across the full speed range despite the absolute energy values spanning nearly two orders of magnitude (265–12,163 J for the CFD planner; Table \ref{tab:Energy_Expenditure}), demonstrating that the planning benefit of wake knowledge is a function of flow geometry rather than flow intensity. The isosurface comparison in Figure \ref{fig:MeanPathEnergy} visualises this structure directly: the difference surface (row 2) reveals the CFD advantage as a ridge deepening monotonically with heading angle at all speeds, while the percentage surface (row 3) confirms that angular dependence is the dominant spatial feature of the planning benefit across the full parameter space.

\begin{figure}[H]
    \centering
    \includegraphics[width=1.0\linewidth]{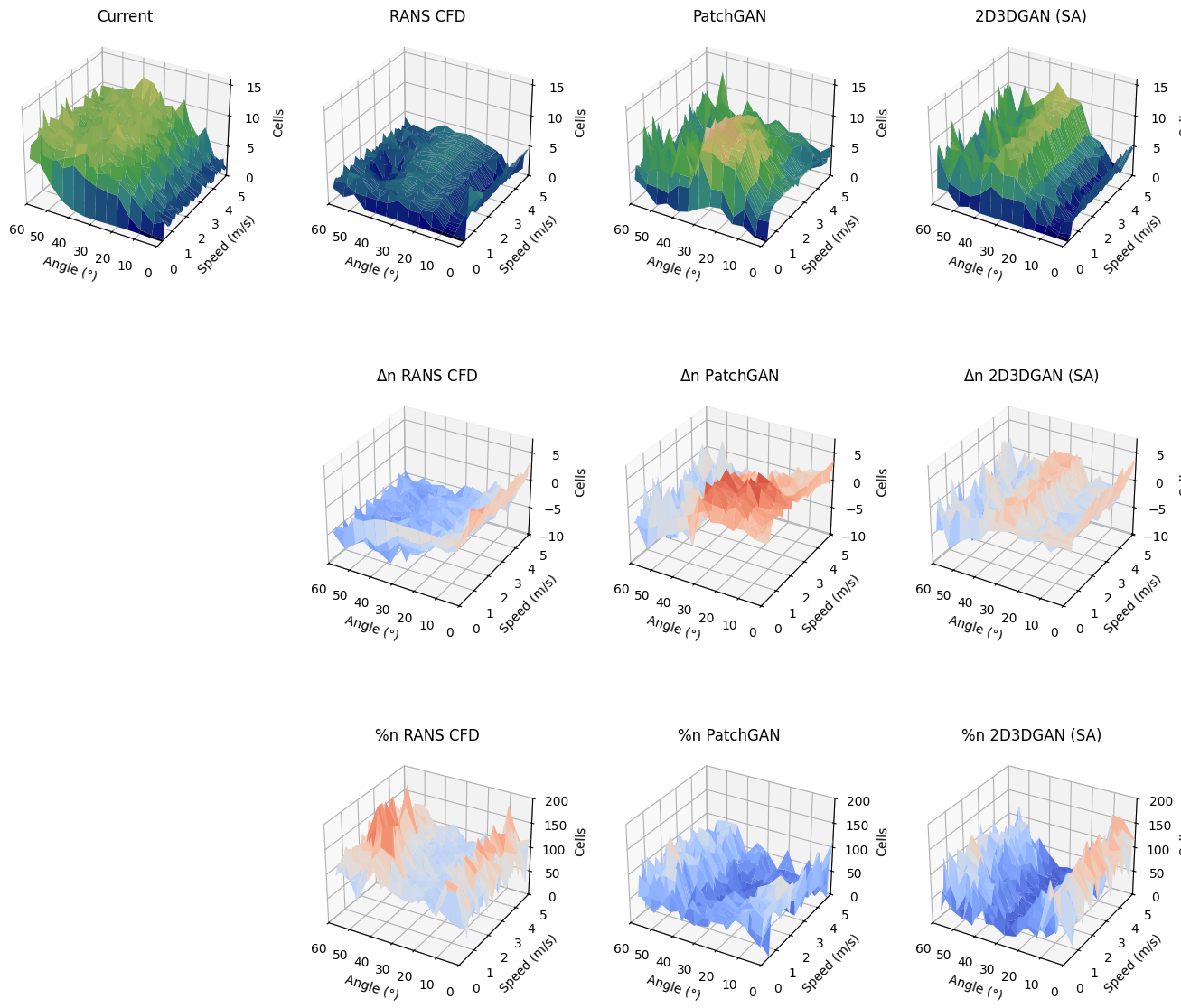}
    \caption{3D isosurface plot of the mean number of high velocity cells encountered for the Current, CFD, PatchGAN and 2D3DGAN (SA) A* planners. Denoted is the angle (x-axis) and the speed (y-axis) of the flow condition that the trajectory was planned on. Vertical (z-axis) used to denote the number of high velocity cells (row 1), change in number of cells (row 2) and the percentage difference (row 3). Differences are given relative to the Current only A* planner.}
    \label{fig:MeanHighVelocityCells}
\end{figure}

\textit{High-velocity cell avoidance}. The reduction in high-velocity encounters achieved by the CFD planner, measured relative to the median-based threshold defined in Equation \ref{eqn:v_threshold}, exhibits a markedly stronger and more condition-dependent pattern than the energy savings (Table \ref{tab:High_Velocity_Cells}; Figure \ref{fig:MeanHighVelocityCells}). At 1.0–2.0 m/s, $45^o - 50^o$, the CFD planner reduces high-velocity cell encounters by 77.83\% relative to the current-only baseline, the largest single-condition benefit in the dataset. At the same speed and $0 - 15^o$ heading, the reduction is only 7.72\%, a factor of approximately ten smaller. This pronounced angular sensitivity reflects the structural difference between the compact, near-axisymmetric propeller race at zero heading, which the current-only planner's direct path avoids with moderate frequency and the spatially extensive, asymmetric separated wake at large heading angles, whose full geometry can only be systematically exploited with explicit spatial knowledge. At intermediate speeds and low angles (0.3–1.0 m/s, $0 - 15^o$), the absolute count of high-velocity encounters is already low for the current-only planner (2.08 ± 3.77 cells; Table \ref{tab:High_Velocity_Cells}), leaving limited margin for further reduction. At the highest speeds and 0–15° heading, the CFD planner achieves 25–31\% reductions, consistent with the elevated wake velocity at higher Reynolds numbers expanding the high-velocity spatial footprint even in straight-ahead conditions.

\begin{figure}[H]
    \centering
    \includegraphics[width=1.0\linewidth]{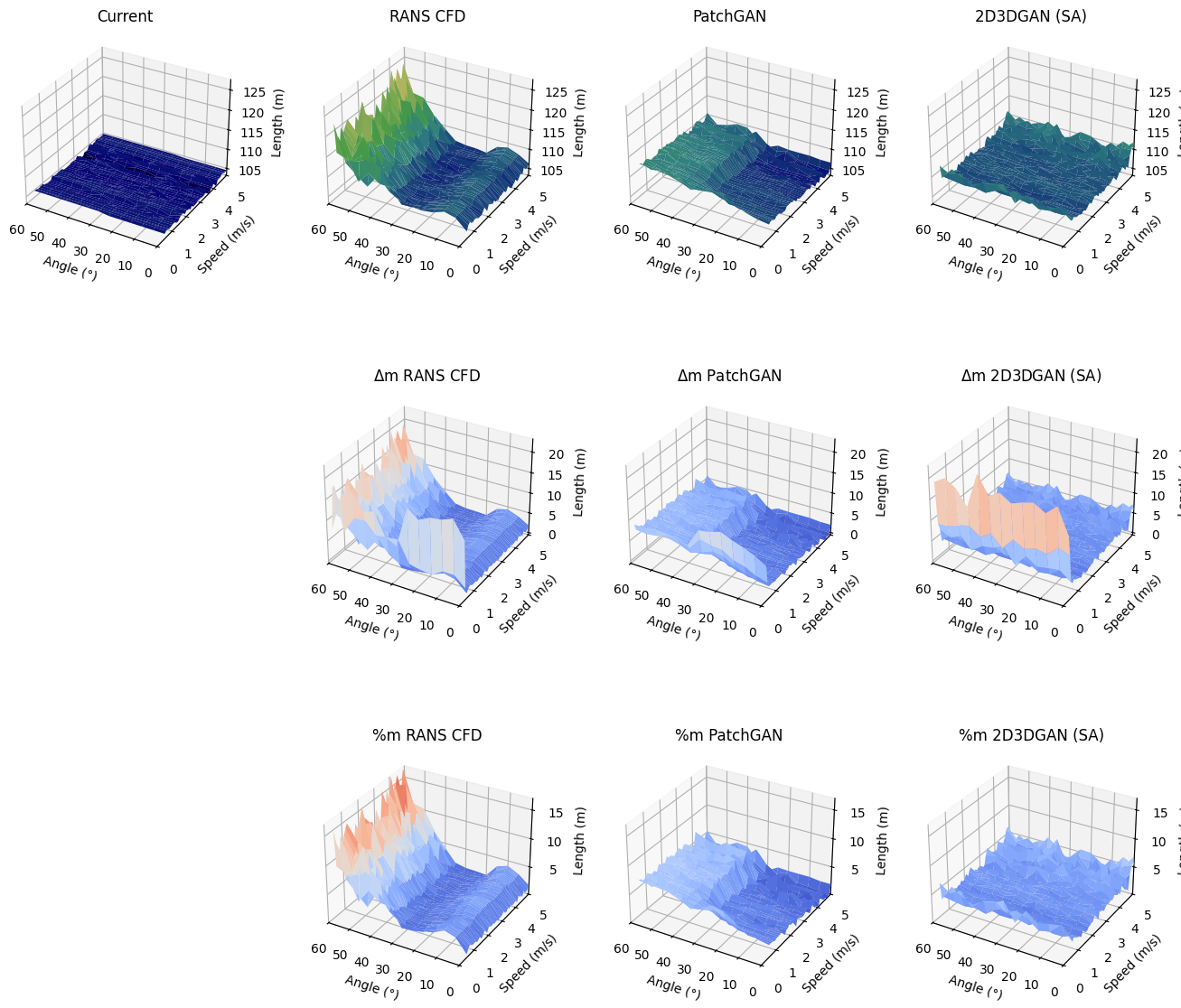}
    \caption{3D isosurface plot of the mean trajectory length for the Current, CFD, PatchGAN and 2D3DGAN (SA) A* planners. Denoted is the angle (x-axis) and the speed (y-axis) of the flow condition that the trajectory was planned on. Vertical (z-axis) used to denote the path length in meters (row 1), change in path length (row 2) and the percentage difference (row 3). Differences are given relative to the Current only A* planner.}
    \label{fig:MeanTrajectoryLength}
\end{figure}

\textit{Path length and the energy–geometry trade-off}. Wake-informed routing incurs a consistent path length extension relative to the current-only planner, which generates the geometrically shortest trajectories in the uniform cost landscape (Table \ref{tab:Path_Length}; Figure \ref{fig:MeanTrajectoryLength}). CFD planner paths are 1.92–8.46\% longer than the current-only baseline, with extensions increasing systematically with heading angle: at $45 - 50^o$, the lateral routing required to circumvent the port-side separated wake compels path length increases of 7.97–8.67\%, whereas at $0 - 15^o$ the more modest avoidance manoeuvre requires only 2.46–2.67\% additional length. The ratio of energy saving to path length extension provides a dimensionless measure of routing efficiency. At 4.0–5.0 m/s, $45 - 50^o$, the CFD planner achieves 12.09\% energy saving at the cost of a 7.97\% path extension, yielding an efficiency ratio of approximately 1.52 — meaning that each additional percent of path length is associated with 1.52\% of energy saving. This ratio is broadly consistent across the high-angle regime (typically 1.4–1.6), indicating that the energy–geometry trade-off is determined primarily by the geometric character of the wake structure rather than by flow speed.

\subsection{Performance of the cGAN Surrogate Planners}
\label{sec:SurrogatePerf}

Energy recovery. Both cGAN planners deliver energy reductions relative to the current-only baseline across most of the parameter space, but their performance is strongly condition-dependent in contrast to the robust and consistent savings of the CFD planner (Table \ref{tab:Energy_Expenditure}). At high heading angles ($45 - 50^o$), both surrogates produce their strongest results: the PatchGAN planner achieves energy savings of 3.56–5.43\% and the 2D3DGAN SA achieves 4.77–6.11\% across this angle range, compared with 11.40–12.51\% for the CFD planner. At the most energetically demanding condition (4.0–5.0 m/s, $45 - 50^o$), the PatchGAN planner recovers 44.9\% and the 2D3DGAN SA recovers 47.1\% of the maximum achievable CFD energy benefit (Table \ref{tab:BenefitOfCFD2}).

At low heading angles ($0 - 15^o$), both surrogates deliver substantially reduced benefits, and the PatchGAN planner records marginal energy degradation relative to the current-only baseline at several $15 - 30^o$ and $30 - 45^o$ bins (0.27–0.68\% higher mean energy; Table \ref{tab:Energy_Expenditure}). The 2D3DGAN SA also records a slight energy increase of 0.28\% at 0.3–1.0 m/s, $0 - 15^o$. These small degradations, while limited in absolute magnitude, indicate that neither surrogate fully resolves the compact near-axisymmetric propeller race at low heading angles, and that both can misdirect the planner relative to the uninformed baseline in these conditions. The PatchGAN's degradation is more frequent and more consistent, in keeping with its documented failure to reliably reproduce the compact wake core at low angles \cite{Zac_PhD_Thesis}.

\textit{High-velocity cell avoidance}. The high-velocity cell metric reveals a more architecturally differentiated and operationally significant performance pattern than energy expenditure (Table \ref{tab:High_Velocity_Cells}). At high heading angles ($45 - 50^o$), both surrogates substantially reduce high-velocity encounters: the PatchGAN planner achieves reductions of 41.51–46.56\% and the 2D3DGAN SA achieves 28.84–46.67\% at these conditions, recovering 59.5\% and 53.8\% of the CFD avoidance benefit respectively at 4.0–5.0 m/s, $45 - 50^o$ (Table \ref{tab:BenefitOfCFD2}).

At low heading angles ($0 - 15^o$), the two architectures diverge sharply. The PatchGAN planner records substantially elevated high-velocity cell counts at several low-angle bins: at 1.0–2.0 m/s, $0 - 15^o$, it traverses 121.05\% more high-velocity cells than the uninformed current-only planner. This represents a complete inversion of the expected performance ordering — the surrogate-informed planner performing substantially worse than the naive baseline on the primary safety metric at these conditions. The mechanism is consistent with the PatchGAN's known failure mode: systematic under-prediction of peak velocity in the compact propeller race causes the planner to route into regions it incorrectly scores as low-cost, producing paths that encounter the wake core more frequently than the direct current-only trajectory. The 2D3DGAN SA does not exhibit this failure to the same degree. At 1.0–2.0 m/s, $0 - 15^o$, it reduces high-velocity encounters by 19.93\% relative to the current-only baseline, maintaining a positive planning benefit where the PatchGAN fails severely. At the low-speed, low-angle extreme (0.3–1.0 m/s, $0 - 15^o$), the 2D3DGAN SA achieves a 43.11\% reduction in high-velocity cells while the PatchGAN registers a 57.46\% increase (Table \ref{tab:High_Velocity_Cells}).

This architectural divergence in high-velocity cell performance at low heading angles constitutes the clearest evidence in these results that standard volumetric reconstruction metrics are insufficient predictors of downstream planning utility. Despite the 2D3DGAN SA's substantially lower MSE ($8.38 \times 10^{-6}$ vs $1.22 \times 10^{-4}$) and higher SSIM (0.992 vs 0.941) relative to the PatchGAN \cite{Zac_PhD_Thesis}, the PatchGAN's specific failure in the compact low-angle wake core has a disproportionately large impact on planning performance because the planner is routed directly into the region it misrepresents. The 2D3DGAN SA's superior global reconstruction fidelity is evidently sufficient to preserve the spatial accuracy of the compact wake structure at these conditions, while the PatchGAN's weaker mean reconstruction breaks down precisely in the high-gradient propeller race region that is most planning-relevant at low angles.

\subsection{Architecture Comparison}
\label{sec:ArchComp}

Across the full parameter space, the PatchGAN GradNorm and 2D3DGAN SA planners show broadly comparable energy performance, with inter-architecture differences of less than one percentage point in most speed–angle bins (Table \ref{tab:Energy_Expenditure}). The 2D3DGAN SA holds a consistent marginal energy advantage at high heading angles with 5.70\% versus 5.43\% at 4.0–5.0 m/s, $45 - 50^o$, consistent with its superior reconstruction of the asymmetric, spatially extended wake structures characteristic of turning conditions. Path length extensions (Table \ref{tab:Path_Length}) are similar for both architectures, though the 2D3DGAN SA consistently extends paths marginally more at high angles, suggesting a slightly more aggressive lateral routing strategy relative to the predicted wake boundary. In neither case do the GAN planners approach the path length extensions of the CFD planner, confirming that neither surrogate fully reproduces the spatial extent of the true wake boundary.

The two NN approximators from prior work \cite{ZacPlanner} being the Current\_NN and CFD\_NN consistently produce inferior trajectories to all four A* variants across both the energy and high-velocity metrics for all conditions (Tables \ref{tab:Energy_Expenditure} and \ref{tab:High_Velocity_Cells}). The CFD\_NN, trained to replicate wake-informed A* trajectories, records energy expenditures 9.96–13.22\% above the current-only A* baseline (Table \ref{tab:Energy_Expenditure}), rendering it worse than the uninformed A* planner on this metric at all conditions. It also produces paths 9.71–14.03\% longer than the current-only A* baseline (Table \ref{tab:Path_Length}) which are substantially longer than any A* variant. This degradation reflects the fundamental limitation of feedforward trajectory regression in high-dimensional, spatially structured flow environments: the network captures statistical regularities in the distribution of A* trajectories but cannot reproduce the condition-specific spatial reasoning that produces near-optimal routing through a complex, spatially heterogeneous cost field. The NN results establish an important negative: the speed advantage of end-to-end trajectory regression is not achievable without a substantial and operationally meaningful loss of path quality, even when the network is trained directly on optimal A* trajectories.

\subsection{Turbulent Cell Traversal Trade-off}
\label{sec:CellTraversal}

All four wake-informed A* planners increase turbulent cell traversals relative to the current-only baseline across the entire parameter space (Table \ref{tab:Turbulent_Cells}). This is not an artefact of the GAN surrogates but is equally pronounced for the ground-truth CFD planner: at 0.3–1.0 m/s, $0 - 15^o$, the CFD planner traverses 52.17\% more turbulent cells than the current-only baseline; the PatchGAN and 2D3DGAN SA show increases of 41.86\% and 27.85\% respectively at the same condition. Both GAN planners exhibit elevated turbulent cell counts across most bins, typically 15–55\% above the current-only baseline, broadly tracking the CFD planner's pattern.

This consistent increase is a direct consequence of the energy-optimal avoidance strategy. In routing around the high-velocity wake core, wake-informed planners guide the AUV through the surrounding shear layers, where the RANS-resolved velocity field exhibits spatially distributed but lower-magnitude velocity fluctuations. The current-only planner, following the geometrically direct path, traverses the coherent wake core and paradoxically spends less time in these peripheral transition zones. The turbulent cell metric therefore captures a genuinely complementary characteristic to the high-velocity metric: whereas the latter measures exposure to peak-energy wake core regions, the former captures the spatial frequency of velocity transitions that excite pitch, yaw, and roll perturbations in the vehicle's attitude control system. A vehicle following a wake-informed trajectory will experience fewer severe, transient hydrodynamic loads than one following the direct path, but will encounter a more prolonged sequence of distributed, lower-amplitude perturbations that cumulatively challenge guidance, navigation and control system performance.

The trade-off between these two operational characteristics decreases in magnitude with heading angle. At $45 - 50^o$, the turbulent cell increase associated with CFD-informed routing is 19–37\% relative to the current-only baseline, compared with 37–59\% at $0 - 15^o$. This improvement at large angles reflects the geometric structure of the asymmetric wake: the strong port-side vortical structures at high heading angles create spatially extended corridors of genuinely lower turbulence intensity alongside the high-velocity separated region, through which the wake-informed planner routes more efficiently than it can at low angles where the wake is compact and the surrounding shear region is comparatively narrow.

\subsection{Computational Performance}
\label{sec:CompPerf}

A* search times across all four planner variants and both NN approximators are reported in Table \ref{tab:Compute_Time}. The current-only planner exhibits the highest and most consistent mean search times (65.2–69.5 s across all speed–angle bins), reflecting the near-uniform cost landscape that requires A* to expand a large fraction of the $128^3$ graph before converging to the optimal path. The RANS CFD planner is substantially faster at high heading angles such as 43.27 ± 25.10 s at 4.0–5.0 m/s, $45 - 50^o$ versus 65.51 ± 19.83 s for the current-only planner, a 34\% reduction, but shows comparable or slightly elevated search times at low heading angles, where the sharp spatial gradients of the compact near-symmetric RANS wake appear to create a locally complex cost landscape that prolongs graph exploration. Notably, the CFD planner exhibits substantially elevated standard deviations at low speeds (e.g., 80.78 ± 597.65 s at 0.3–1.0 m/s, $0 - 15^o$), indicating occasional pathological search behaviour in conditions where the compact, localised wake structure generates narrow cost-surface features that A* explores inefficiently from certain start positions.

Both cGAN planners are consistently faster at A* search than the RANS CFD planner across all conditions, and faster than the current-only planner at high heading angles. At 4.0–5.0 m/s, $45 - 50^o$, the PatchGAN planner completes search in 37.36 ± 18.03 s and the 2D3DGAN SA in 38.69 ± 16.16 s, compared with 43.27 s for CFD and 65.51 s for the current-only planner. This search-time advantage is not attributable to cGAN inference speed, which contributes less than 0.1\% of total planning time in both architectures, but rather to the structural character of the predicted cost surface. The smoother wake boundaries characteristic of GAN-predicted fields produce a more regularly navigable cost landscape, and the substantially lower standard deviations of GAN search times relative to CFD further indicate that the surrogate's representation of the wake does not generate the narrow, high-gradient cost features responsible for the CFD planner's occasional pathological search behaviour. The GAN-informed planners therefore offer a dual computational advantage: near-instantaneous field generation from scalar inputs in place of multi-hour CFD simulation, and more efficient and predictable A* search execution relative to the ground-truth field.

The two NN approximators operate in an entirely different computational regime, producing trajectories in under 0.40 ms regardless of flow condition. This is a reduction of approximately five orders of magnitude relative to all A* variants (Table \ref{tab:Compute_Time}). This speed advantage is structurally decoupled from the quality of the environmental model: the networks generate trajectories in a single feedforward pass from scalar metadata, without constructing or querying a flow field. The NN and cGAN-informed A* approaches therefore occupy distinct positions in the computation–quality trade-off space: the cGAN planners enable a substantial quality improvement within a search-based framework while remaining computationally tractable, whereas the NN approximators sacrifice a significant and measurable degree of trajectory optimality to eliminate the search process entirely.

\section{Discussion}
\label{sec:Discussion}

The results of this benchmark, taken together, permit a specific and operationally meaningful answer to the central question of this work: how much does wake knowledge matter for AUV path planning, and how much of its value do cGAN surrogates recover? This discussion quantifies that answer, interprets the trade-offs the data reveal, situates the findings within the broader challenge of surrogate-enabled path planning, and acknowledges the principal limitations of the study.

\subsection{The Value of Hydrodynamic Information}
\label{sec:DiscussionHydroInformation}

The clearest finding of this work is the confirmation that detailed, 3D wake knowledge has substantial and quantifiable operational value for AUV path planning in the LAR scenario. This is a value that cannot be captured by a uniform current model and that grows with the severity of the flow condition. Table \ref{tab:BenefitOfCFD} summarises the performance gap between the Current and CFD planners, which represents the maximum available benefit of perfect hydrodynamic information.

\begin{table}[!ht]
    \centering
    \setlength{\tabcolsep}{6pt}
    \begin{tabular}{cc|c|c|c|c} \hline \hline
    Speed & Angle &
    Energy saving & Path length increase &
    High-vel.\ cells saved & Turb.\ cells increase \\ \hline \hline
    \multirow{2}{*}{0.3 -- 1.0}
        & 0 -- 15  & 7.3\%  & $+$4.3\% & 19.2\% & $+$52.2\% \\
        & 45 -- 50 & 12.5\% & $+$8.5\% & 68.1\% & $+$37.3\% \\ \hline
    \multirow{2}{*}{1.0 -- 2.0}
        & 0 -- 15  & 7.1\%  & $+$2.5\% &  7.7\% & $+$58.8\% \\
        & 45 -- 50 & 11.8\% & $+$8.2\% & 77.8\% & $+$28.6\% \\ \hline
    \multirow{2}{*}{2.0 -- 4.0}
        & 0 -- 15  & 7.1\%  & $+$2.6\% & 29.0\% & $+$50.0\% \\
        & 45 -- 50 & 11.7\% & $+$8.4\% & 73.4\% & $+$25.6\% \\ \hline
    \multirow{2}{*}{4.0 -- 5.0}
        & 0 -- 15  & 7.1\%  & $+$2.7\% & 25.2\% & $+$41.0\% \\
        & 45 -- 50 & 12.1\% & $+$8.0\% & 69.8\% & $+$19.2\% \\ \hline \hline
    \end{tabular}
    \caption{Performance benefit of full RANS CFD wake knowledge relative to the
    current-only baseline, across eight representative speed--angle bins. Energy
    saving and high-velocity cell reduction are positive when the CFD planner
    outperforms the current-only planner; path length increase and turbulent cell
    increase are positive for all wake-informed planners as a consequence of routing
    around the wake core. The 2.0--4.0~m/s bin is the arithmetic mean of the
    2.0--3.0 and 3.0--4.0 sub-bins reported in Tables~\ref{tab:Energy_Expenditure}--\ref{tab:Turbulent_Cells}.}
    \label{tab:BenefitOfCFD}
\end{table}

The energy savings from wake-aware planning are primarily driven by heading angle, with only weak dependence on operating speed. At near-dead-ahead conditions (0--15$^\circ$), savings stabilise in the narrow range of 7.1--7.3\% across all five speed bins, demonstrating that the navigational benefit of complete wake knowledge is determined by wake geometry rather than Reynolds number over the tested range. As heading angle increases to 45--50$^\circ$, savings rise consistently to 11.75--12.51\%, regardless of speed. The overall range across all twenty speed--angle bins spans from 5.74\% at 1.0--2.0 m/s, 15--30$^\circ$ to 12.51\% at 0.3--1.0 m/s, 45--50$^\circ$. This pattern is physically consistent: at near-zero heading angles the XLUUV wake is near-axisymmetric and comparatively compact, limiting the navigational value of knowing its spatial structure. As heading angle grows, the wake becomes increasingly asymmetric and dominated by stronger shear layers, tip vortices, and separated flow on the leeward side of the hull. These structures whose spatial extent and energetic cost can only be systematically exploited with explicit volumetric knowledge. The benefit of wake information therefore scales with wake asymmetry rather than flow intensity.

The high-velocity cell avoidance data reveals an even sharper dependence on operating regime. At the most demanding conditions (1.0–2.0 m/s, 45–50 degrees), the CFD-informed planner reduces encounters with the high-velocity wake core by 77.8\% compared to the naive baseline. This finding has implications beyond energy efficiency: high-velocity regions in the propeller wake represent the primary source of control saturation risk during LAR, where localised forces can exceed the vehicle's thruster capacity. The ability of the CFD planner to reduce encounters with these regions, so substantially and consistently across conditions, demonstrates that wake-aware planning provides a structural safety benefit, not merely an energy optimisation benefit, that should factor into mission planning for any AUV operating at close range to a larger platform.

The turbulent cell metric introduces the most nuanced finding of the study. All wake-informed planners: CFD, PatchGAN, and 2D3DGAN SA, consistently increase turbulent cell traversals relative to the naive baseline, by between 14\% and over 50\% depending on the condition. This is not a failure of the planners but a direct consequence of their success in avoiding the high-velocity core: in doing so, they route the vehicle through the surrounding shear layers, where lower-magnitude but spatially distributed velocity fluctuations are classified as turbulent under the no-threshold counting metric used. The current planner, taking the shortest direct route, cuts through the coherent wake core and paradoxically spends less time in these peripheral turbulent zones. This finding has an important implication for AUV control system design: energy-aware wake routing shifts the nature of the disturbance environment rather than eliminating it. A vehicle following a CFD-informed trajectory will face less extreme, but more prolonged and spatially distributed, perturbations than one following a naive shortest-path trajectory. Future controller designs for LAR operations should account for this trade-off explicitly.

\subsection{Proximity to CFD with Surrogate Information}
\label{sec:DiscussionOptimality}

The central practical question of this work is whether the cGAN-predicted fields are sufficient substitutes for ground-truth CFD data in the path planning loop. Table~\ref{tab:BenefitOfCFD2} answers this by expressing GAN planner performance as the fraction of the maximum available CFD benefit that each surrogate recovers, evaluated across the energy and high-velocity cell metrics at the most demanding operating conditions (4.0--5.0 m/s, 45--50 degrees).

Both cGAN planners recover approximately 45--60\% of the CFD energy benefit and high-velocity cell avoidance benefit at the most demanding conditions, while closing a smaller but still substantial fraction of the path length and turbulence trade-offs. In absolute terms at 4.0--5.0 m/s, 45--50$^\circ$, the GAN planners reduce energy expenditure by 5.4--5.7\% relative to the naive baseline (compared to the CFD planner's 12.1\%), and reduce high-velocity cell encounters by 37.6--41.6\% in absolute terms (compared to the CFD planner's 69.8\%), recovering 53.8--59.5\% of the CFD avoidance benefit (Table~\ref{tab:BenefitOfCFD2}). A vehicle equipped with either GAN planner is substantially safer and more efficient than one operating on uniform current data alone, across essentially all operating conditions.

\begin{table}[!ht]
    \centering
    \begin{tabular}{c|c|c|c} \hline \hline
    Metric & CFD benefit & PatchGAN recovery & 2D3DGAN SA recovery \\ \hline \hline
    Energy saving & 12.1\% & 44.9\% of CFD benefit & 47.1\% of CFD benefit \\
    High-vel cell reduction & 69.8\% & 59.5\% of CFD benefit  & 53.8\% of CFD benefit    \\
    Path length increase & +8.0\% & 49.3\% of CFD benefit  & 46.3\% of CFD benefit    \\
    Turbulent cells (worst case) & +19\% & 74.2\% of CFD benefit  & 86.8\% of CFD benefit    \\  \hline
    \end{tabular}
    \caption{Fraction of maximum CFD knowledge recovered by cGAN surrogate architectures (4.0 - 5.0 m/s, 45-50 degrees)}
    \label{tab:BenefitOfCFD2}
\end{table}

The performance difference between the two cGAN architectures is smaller than their reconstruction metric gap might suggest at high heading angles, but is pronounced and directionally consistent at low heading angles, where the architectures diverge in a practically important way. The 2D3DGAN~SA, which achieves markedly superior SSIM (0.992 vs 0.941) and MSE ($8.38 \times 10^{-6}$ vs $1.22 \times 10^{-4}$), holds a marginal energy advantage at high angles of attack (5.70\% vs 5.43\% saving at 4.0--5.0 m/s, 45--50 degrees) consistent with its superior reconstruction of the asymmetric, spatially extended wake at turning conditions. At low heading angles, the 2D3DGAN~SA maintains positive high-velocity cell avoidance performance (19.93\% reduction relative to current-only at 1.0--2.0 m/s, 0--15$^\circ$; 43.11\% reduction at 0.3--1.0 m/s, 0--15$^\circ$). The PatchGAN, by contrast, fails severely in this regime: at 1.0--2.0 m/s, 0--15$^\circ$, it traverses 121.05\% more high-velocity cells than the uninformed current-only baseline, representing a complete inversion of the expected performance ordering. The mechanism is the PatchGAN's documented under-prediction of peak velocity in the compact, near-axisymmetric propeller race at low heading angles \cite{Zac_PhD_Thesis}, which causes the planner to route into the wake core it misidentifies as low-cost. The 2D3DGAN~SA is therefore the safer and more consistent planning surrogate across the full operating envelope: it matches or exceeds the PatchGAN on energy expenditure at all conditions, maintains positive high-velocity avoidance where the PatchGAN catastrophically fails, and yields only a marginal high-velocity recovery concession at high heading angles (53.8\% vs 59.5\% of the CFD benefit at 4.0--5.0 m/s, 45--50$^\circ$). Its primary disadvantage relative to the PatchGAN is computational (25.7~GB of GPU memory versus 1.21~GB) rather than planning-metric performance.

The two NN approximators from prior work \cite{ZacPlanner} occupy a qualitatively different position in the computation--quality trade-off space. Both produce trajectories in under 0.4~ms, but at a substantial cost to path quality: the CFD\_NN records energy expenditures 9.96--13.22\% above the current-only A* baseline, rendering it worse than the uninformed A* planner on this metric at all conditions. This establishes an important negative result: the speed advantage of end-to-end trajectory regression is not achievable without a measurable and operationally significant loss of path quality, even when the network is trained directly on optimal A* trajectories. The cGAN planners and the NN approximators therefore address different operational constraints and should be understood as complementary options on the deployment spectrum, not as competing solutions.

A notable secondary finding concerns the A* search time advantage of the GAN planners over the CFD planner. Both GAN-informed A* planners are faster than the CFD-informed A* planner at all conditions and substantially faster at high heading angles: the PatchGAN planner reduces A* search time by up to 76.5\% relative to the current-only baseline at 45--50$^\circ$ (Table~\ref{tab:Compute_Time}). This is not a direct consequence of the GAN field quality, but of the cost landscape structure: the smoother wake boundaries in the GAN fields produce a more regular cost surface that A* explores more efficiently than the sharp gradients of the RANS CFD field. This is an incidental but practically relevant benefit — the GAN-informed planner is not only cheaper to set up (surrogate inference vs CFD simulation) but also cheaper to run at planning time.

\subsection{Reconstruction Fidelity Versus Downstream Task Performance}
\label{sec:DiscussionDownstreamTasks}

The results of this work contribute a concrete data point to the broader and under-explored question of when ML surrogate reconstruction accuracy is sufficient for downstream engineering tasks. In the surrogate modelling literature, it is common to evaluate models solely on reconstruction metrics such as MSE, PSNR, and SSIM, implicitly assuming that a more accurate field will produce correspondingly better downstream task performance. The present benchmark reveals a more specific finding: global reconstruction metrics substantially underestimate the planning performance gap when reconstruction errors are spatially concentrated in energetically critical regions. The PatchGAN's global MSE of $1.22 \times 10^{-4}$ is approximately 15 times larger than the 2D3DGAN~SA's $8.38 \times 10^{-6}$, and its SSIM of 0.941 is modestly below the 2D3DGAN~SA's 0.992, a meaningful but not extreme gap. Yet at 1.0--2.0 m/s, $0 - 15^\circ$, these moderate aggregate differences translate into a planning performance divergence of 121.05 percentage points on the high-velocity cell metric: the PatchGAN routes the vehicle into 121.05\% more high-velocity cells than the uninformed current-only baseline, while the 2D3DGAN~SA reduces them by 19.93\%. Global volume-level metrics, dominated by the large ambient-flow region where both architectures perform well, do not capture whether reconstruction errors are spatially concentrated in the planning-relevant wake core. The PatchGAN's failure is not visible in its aggregate reconstruction score, but it is operationally catastrophic precisely because those errors occur in the small spatial region that determines path safety at low heading angles.

This finding has implications for how future surrogate models for path planning applications should be designed and evaluated. Optimising reconstruction metrics against the full volume including the large ambient-flow region where all architectures perform well may not be the correct training objective for planning-relevant surrogates. A targeted training objective that applies elevated weight to the energetically significant wake region, such as a spatially resolved version of the GMSE loss \cite{ZacLoss} applied specifically to voxels within a defined distance of the XLUUV hull and propeller disk, may better align the surrogate's reconstruction quality with its planning utility. Alternatively, end-to-end training of the surrogate within the planning loop minimising planning cost directly rather than field reconstruction error represents a more principled but computationally demanding approach that warrants future investigation.

\subsection{Limitations}
\label{sec:DiscussionLimitations}

\subsubsection{The Pseudo-Static Assumption}

The most significant methodological limitation of this work is the pseudo-static treatment of the flow field: for each planned trajectory, the XLUUV speed and heading are held constant and the corresponding RANS mean flow field is used throughout. This is a deliberate and physically motivated simplification as the RANS solutions capture the time-averaged wake structure that governs the energy and safety characteristics of the LAR approach, and the manoeuvre duration is short relative to the timescale of mean wake evolution. However, real LAR operations involve a host platform that may manoeuvre during the approach, and the instantaneous flow field exhibits turbulent fluctuations around the RANS mean that are not represented in the dataset.

There are two distinct paths toward addressing this limitation. The first is the incorporation of unsteady RANS (URANS) or Large Eddy Simulation (LES) data, which would provide time-resolved velocity fields and allow the planner to account for temporal evolution of the wake. Training cGAN surrogates on such data would require an extension of the current architecture to handle spatio-temporal sequences, for instance via ConvLSTM-augmented generators or 4D convolutional architectures \cite{StridedConvols}. The additional temporal dimension would substantially increase data generation costs and architectural complexity, but would yield surrogates capable of producing time-varying flow field predictions that are directly usable in a receding-horizon planning framework. The second path is the integration of online flow sensing including acoustic Doppler current profilers, pressure sensor arrays, or wake-following sonar, to update the surrogate's conditional inputs during the approach, enabling the planner to track real-time deviations from the mean flow model. Both directions are tractable extensions of the pipeline established in this work and represent the most important avenue for future development.

\subsubsection{Geometric Generalisation}

The dataset and all models in this work are specific to the generalised XLUUV geometry described in Section \ref{sec:VehicleGeometry}. While this geometry is intentionally designed to be representative of the XLUUV class rather than specific to a single platform \cite{WAKESET}, the trained cGAN models cannot be expected to generalise to host vehicles of substantially different form. For example, this could be platforms with different length-to-diameter ratios, propeller configurations, or hull appendages. Extending the surrogate to a family of host geometries would require either a dataset that spans geometry as a conditioning variable, or a domain-adaptive training strategy that transfers the learned flow structure prior from the reference geometry to new configurations. This is a significant but tractable extension, particularly given that the metadata conditioning architecture already in place for speed and heading could, in principle, be extended to include geometric parameters.

\subsubsection{AUV Kinodynamic Constraints}

The A* planner used in this work optimises energy expenditure subject only to the hydrodynamic cost field and the discrete 26-connected voxel adjacency structure. It does not enforce the kinodynamic constraints of a real AUV such as minimum turning radius, maximum angular rate, thrust saturation, or the coupled interaction between vehicle motion and the evolving flow field. The resulting trajectories are therefore geometrically and energetically optimal within the voxel grid but may not be directly executable by a specific vehicle without post-processing for dynamic feasibility. The integration of kinodynamic constraints into the planning framework, via motion primitive libraries or trajectory optimisation post-processing, is a necessary step toward operational deployment and is identified as a priority direction for future work \cite{ZacPlanner}.

\subsubsection{Toward Real-Time Onboard Deployment}

Despite the limitations above, the pipeline demonstrated here represents a concrete and near-term route toward wake-aware path planning that is computationally tractable for onboard use. The remaining gap between current GAN surrogate performance (recovering 45–60\% of the CFD planning benefit) and the theoretical optimum (full CFD knowledge) should be understood not as a failure but as a lower bound on what is achievable with a single inference, fixed-weight surrogate operating from scalar metadata alone. Each of the limitations identified above: pseudo-static flow, geometric specificity, no kinodynamic constraints — represents a dimension along which the pipeline can be progressively improved without altering its fundamental architecture. The hierarchical 2D→3D pipeline, the energy-weighted A* framework, and the four-metric evaluation suite established in this work provide the benchmark infrastructure against which those improvements can be quantitatively assessed.

\section{Conclusion}
\label{sec:Conclusion}

This work has addressed a specific and practically important gap in autonomous underwater vehicle path planning: the absence of any systematic, quantitative assessment of what is lost by ignoring three-dimensional wake structure during close-proximity operations, and whether a computationally tractable surrogate model can recover those losses. By benchmarking four environmental knowledge levels: uniform current, ground-truth RANS CFD, and two cGAN architectures across 19,800 independently generated trajectories spanning 550 flow conditions, we have produced the first such assessment for the AUV launch and recovery scenario.

Three findings define the contribution of this work. First, wake knowledge has substantial, quantifiable operational value that scales with operating severity. Relative to planning on a uniform current model, a planner with access to full CFD wake data achieves energy savings of 5.7–12.5\%, reduces encounters with high-velocity flow regions by up to 77.8\%, and generates these benefits at the cost of longer, more tortuous trajectories through the wake periphery. These are not incremental improvements but operationally meaningful differences that compound over repeated mission cycles and matter acutely at the moments of greatest control risk — high-speed, high-heading-angle LAR approaches where the propeller race and separated wake dominate the near-field environment.

Second, cGAN surrogates recover a substantial and operationally useful fraction of this benefit — approximately 45–60\% of the CFD energy saving and high-velocity cell avoidance advantage — at inference times that are orders of magnitude lower than RANS CFD simulation. The PatchGAN GradNorm pipeline, with an end-to-end inference time of approximately 28 $\mu s$ from scalar operating condition inputs to a populated 1283 voxel planning grid, is deployable on hardware in the class of modern embedded GPU accelerators currently in use in advanced AUV payloads. The 2D3DGAN SA pipeline achieves marginally higher energy performance at high heading angles but requires 25.7 GB of GPU memory for the 3D generator alone, placing it beyond current embedded hardware limits. This distinction defines a clear near-term roadmap: the PatchGAN path is deployable now; the 2D3DGAN SA path is appropriate for pre-deployment or surface-assisted planning in the near term, and for onboard deployment as embedded GPU capability continues to advance.

Third, the reconstruction fidelity of a cGAN surrogate as measured by standard metrics including MSE, PSNR, and SSIM does not straightforwardly predict its utility as a planning environment. The PatchGAN GradNorm, despite achieving inferior global reconstruction metrics, performs comparably on energy expenditure but fails critically at low heading angles on the safety metric, where its specific failure mode: systematic under-prediction of peak velocity in the compact propeller race which routes the planner into 121\% more high-velocity cells than the naive baseline. The 2D3DGAN SA's superior reconstruction fidelity is sufficient to avoid this failure, demonstrating that global reconstruction metrics substantially underestimate the planning consequence of spatially localised reconstruction errors. This finding contributes a concrete and domain-specific data point to the broader surrogate modelling literature on the relationship between reconstruction accuracy and downstream task performance, and motivates the development of planning-relevant training objectives for flow field surrogates.

The principal limitation of this work is the pseudo-static treatment of the flow environment, which is physically motivated for the LAR scenario but does not account for host vehicle manoeuvring or unsteady turbulent fluctuations around the RANS mean. Addressing this limitation through unsteady simulation data, temporal surrogate architectures, or online flow sensing represents the most important direction for future development. Additional directions include the integration of AUV kinodynamic constraints into the planning framework, the extension of the geometric dataset to a family of host vehicle configurations, and the investigation of planning-targeted training objectives for the surrogate network.

The hierarchical cGAN prediction pipeline, the four-metric evaluation framework, and the 19,800-trajectory benchmark established in this work together provide the methodological infrastructure for these future developments and set a quantitative baseline against which they can be measured. The core conclusion is straightforward: knowing the 3D wake matters significantly for AUV path planning in close-proximity operations, and cGAN surrogates make it tractable to act on that knowledge in real time. The cost of not knowing in energy expenditure, in safety margin, and in mission endurance is now quantified.

\section*{Acknowledgements}

The dataset used here to develop the comparison, as well as the trained models, was assisted by the resources of the Australian National Computational Infrastructure (NCI) and GADI Supercomputer which is supported by the Australian government. This work was conducted as part of the Australian HPC-AI Talent Program. 

\bibliographystyle{unsrturl}
\bibliography{refs}

\clearpage 

\appendix

\section{Appendix A: Full Metric Tables}
\label{App:A}

\begin{sidewaystable}[]
\centering
\scriptsize
\begin{tabular}{lll|l|ll|ll|ll|ll|ll}
\hline \hline
Metric & Speed    & Angle   & Current                & CFD                    & \%      & PatchGAN               & \%      & 2D3DGAN (SA)               & \%      & CFD\_NN                & \%      & Current\_NN      & \%      \\ \hline \hline
Energy & 0.3 - 1.0 & 0 - 15  & 285.81 $\pm$ 192.94    & 265.77 $\pm$ 183.35    & 7.27$\downarrow$  & 281.42 $\pm$ 189.25    & 1.55$\downarrow$  & 286.61 $\pm$ 190.50    & 0.28$\uparrow$   & 284.04 $\pm$ 192.82 & 0.62$\downarrow$  & 322.65 $\pm$ 238.97 & 12.11$\uparrow$  \\
$E$ [$J$] [$ \downarrow $]       &           & 15 - 30 & 288.09 $\pm$ 193.52    & 270.28 $\pm$ 185.11    & 6.38$\downarrow$  & 290.06 $\pm$ 192.01    & 0.68$\uparrow$   & 286.17 $\pm$ 189.82    & 0.67$\downarrow$  & 287.53 $\pm$ 194.30 & 0.19$\downarrow$  & 322.12 $\pm$ 237.93 & 11.16$\uparrow$  \\
       &           & 30 - 45 & 293.18 $\pm$ 194.27    & 267.10 $\pm$ 181.84    & 9.31$\downarrow$  & 293.97 $\pm$ 195.24    & 0.27$\uparrow$   & 291.41 $\pm$ 192.26    & 0.61$\downarrow$  & 291.15 $\pm$ 193.23 & 0.70$\downarrow$  & 326.46 $\pm$ 234.75 & 10.74$\uparrow$  \\
       &           & 45 - 50 & 296.80 $\pm$ 193.12    & 261.86 $\pm$ 171.49    & 12.51$\downarrow$ & 281.82 $\pm$ 181.13    & 5.18$\downarrow$  & 281.70 $\pm$ 180.11    & 5.22$\downarrow$  & 286.94 $\pm$ 179.93 & 3.38$\downarrow$  & 323.07 $\pm$ 232.58 & 8.48$\uparrow$   \\
       & 1.0 - 2.0 & 0 - 15  & 1534.55 $\pm$ 575.47   & 1429.44 $\pm$ 548.69   & 7.09$\downarrow$  & 1498.50 $\pm$ 566.37   & 2.38$\downarrow$  & 1500.02 $\pm$ 562.86   & 2.28$\downarrow$  & 1499.89 $\pm$ 572.76 & 2.28$\downarrow$  & 1730.20 $\pm$ 792.48 & 11.99$\uparrow$  \\
       &           & 15 - 30 & 1542.98 $\pm$ 575.43   & 1456.84 $\pm$ 549.77   & 5.74$\downarrow$  & 1537.94 $\pm$ 572.24   & 0.33$\downarrow$  & 1500.74 $\pm$ 562.45   & 2.78$\downarrow$  & 1525.17 $\pm$ 569.47 & 1.16$\downarrow$  & 1731.34 $\pm$ 792.19 & 11.51$\uparrow$  \\
       &           & 30 - 45 & 1544.56 $\pm$ 577.72   & 1425.30 $\pm$ 544.23   & 8.03$\downarrow$  & 1529.36 $\pm$ 568.88   & 0.99$\downarrow$  & 1512.89 $\pm$ 563.80   & 2.07$\downarrow$  & 1520.92 $\pm$ 563.24 & 1.54$\downarrow$  & 1734.02 $\pm$ 775.88 & 11.56$\uparrow$  \\
       &           & 45 - 50 & 1547.69 $\pm$ 578.75   & 1375.97 $\pm$ 524.98   & 11.75$\downarrow$ & 1493.01 $\pm$ 565.60   & 3.60$\downarrow$  & 1465.31 $\pm$ 550.38   & 5.47$\downarrow$  & 1506.58 $\pm$ 591.86 & 2.69$\downarrow$  & 1697.98 $\pm$ 767.39 & 9.26$\uparrow$   \\
       & 2.0 - 3.0 & 0 - 15  & 4045.79 $\pm$ 984.95   & 3770.71 $\pm$ 974.05   & 7.04$\downarrow$  & 3932.06 $\pm$ 961.68   & 2.85$\downarrow$  & 3982.16 $\pm$ 975.14   & 1.59$\downarrow$  & 3981.27 $\pm$ 1018.47 & 1.61$\downarrow$  & 4580.51 $\pm$ 1614.04 & 13.22$\uparrow$  \\
       &           & 15 - 30 & 4078.40 $\pm$ 982.01   & 3838.96 $\pm$ 954.44   & 6.05$\downarrow$  & 4015.82 $\pm$ 960.40   & 1.55$\downarrow$  & 3962.18 $\pm$ 970.27   & 2.89$\downarrow$  & 4054.91 $\pm$ 1006.48 & 0.58$\downarrow$  & 4580.58 $\pm$ 1588.97 & 11.60$\uparrow$  \\
       &           & 30 - 45 & 4088.97 $\pm$ 990.74   & 3783.50 $\pm$ 938.81   & 7.76$\downarrow$  & 4020.81 $\pm$ 972.26   & 1.68$\downarrow$  & 4016.61 $\pm$ 973.49   & 1.79$\downarrow$  & 4051.15 $\pm$ 977.13 & 0.93$\downarrow$  & 4611.35 $\pm$ 1595.08 & 12.01$\uparrow$  \\
       &           & 45 - 50 & 4070.17 $\pm$ 998.40   & 3631.34 $\pm$ 911.55   & 11.40$\downarrow$ & 3927.78 $\pm$ 945.64   & 3.56$\downarrow$  & 3880.72 $\pm$ 954.50   & 4.77$\downarrow$  & 3972.78 $\pm$ 900.42 & 2.42$\downarrow$  & 4493.74 $\pm$ 1489.22 & 9.89$\uparrow$   \\
       & 3.0 - 4.0 & 0 - 15  & 7774.88 $\pm$ 1447.87  & 7238.00 $\pm$ 1462.36  & 7.15$\downarrow$   & 7547.74 $\pm$ 1429.76  & 2.96$\downarrow$  & 7683.14 $\pm$ 1468.04  & 1.19$\downarrow$  & 7767.26 $\pm$ 1544.43 & 0.10$\downarrow$  & 8875.77 $\pm$ 2801.74 & 13.22$\uparrow$  \\
       &           & 15 - 30 & 7844.26 $\pm$ 1454.25  & 7391.47 $\pm$ 1441.51  & 5.94$\downarrow$  & 7646.23 $\pm$ 1384.10  & 2.56$\downarrow$  & 7672.88 $\pm$ 1485.11  & 2.21$\downarrow$  & 7868.11 $\pm$ 1495.99 & 0.30$\uparrow$   & 8914.50 $\pm$ 2796.72 & 12.77$\uparrow$  \\
       &           & 30 - 45 & 7877.99 $\pm$ 1462.64  & 7280.41 $\pm$ 1412.26  & 7.88$\downarrow$  & 7725.92 $\pm$ 1413.79  & 1.95$\downarrow$  & 7745.55 $\pm$ 1446.89  & 1.70$\downarrow$  & 7823.39 $\pm$ 1440.17 & 0.70$\downarrow$  & 8848.91 $\pm$ 2643.56 & 11.61$\uparrow$  \\
       &           & 45 - 50 & 7870.62 $\pm$ 1464.19  & 6985.82 $\pm$ 1385.28  & 11.91$\downarrow$ & 7474.30 $\pm$ 1390.98  & 5.17$\downarrow$  & 7404.10 $\pm$ 1371.04  & 6.11$\downarrow$  & 7807.85 $\pm$ 1465.20 & 0.80$\downarrow$  & 8717.78 $\pm$ 2577.73 & 10.21$\uparrow$  \\
       & 4.0 - 5.0 & 0 - 15  & 12799.09 $\pm$ 2028.82 & 11922.45 $\pm$ 2137.10 & 7.09$\downarrow$   & 12651.61 $\pm$ 1975.53 & 1.16$\downarrow$  & 12716.28 $\pm$ 2174.81 & 0.65$\downarrow$  & 12771.28 $\pm$ 2219.23 & 0.22$\downarrow$  & 14583.44 $\pm$ 4360.70 & 13.03$\uparrow$  \\
       &           & 15 - 30 & 12901.07 $\pm$ 2032.77 & 12163.73 $\pm$ 2014.78 & 5.88$\downarrow$  & 12601.06 $\pm$ 2012.41 & 2.35$\downarrow$  & 12650.76 $\pm$ 2064.11 & 1.96$\downarrow$  & 12862.73 $\pm$ 2069.12 & 0.30$\downarrow$  & 14550.00 $\pm$ 4300.18 & 12.01$\uparrow$  \\
       &           & 30 - 45 & 13001.67 $\pm$ 2068.19 & 12004.95 $\pm$ 2031.64 & 7.97$\downarrow$  & 12512.28 $\pm$ 1981.76 & 3.84$\downarrow$  & 12891.85 $\pm$ 2038.45 & 0.85$\downarrow$  & 12921.89 $\pm$ 1990.77 & 0.62$\downarrow$  & 14681.99 $\pm$ 4200.70 & 12.14$\uparrow$  \\
       &           & 45 - 50 & 12975.85 $\pm$ 2094.28 & 11496.03 $\pm$ 1989.99 & 12.09$\downarrow$ & 12290.08 $\pm$ 2009.78 & 5.43$\downarrow$  & 12257.11 $\pm$ 2057.78 & 5.70$\downarrow$  & 12810.86 $\pm$ 2027.26 & 1.28$\downarrow$  & 14335.59 $\pm$ 3942.32 & 9.96$\uparrow$   \\ \hline \hline
\end{tabular}

\caption{Energy expenditure (\ref{eqn:PathEnergy}) in joules denoted by $E$ for the A* planners and neural networks. Mean and standard deviation provided for the current‐informed A* path planner (Current), CFD informed, PatchGAN informed and 2D3DGAN informed A* planners. Also provided are the CFD\_NN and Current\_NN neural network approximations. Arrows indicate whether each method’s mean is higher (↑) or lower (↓) than the Current‐informed A* planner.}
\label{tab:Energy_Expenditure}
\end{sidewaystable}

\begin{sidewaystable}[]
\centering
\scriptsize
\begin{tabular}{lll|l|ll|ll|ll|ll|ll}
\hline \hline
Metric      & Speed     & Angle   & Current            & CFD                & \%         & PatchGAN             & \%         & 2D3DGAN (SA)            & \%         & CFD\_NN             & \%            & Current\_NN             & \%            \\
\hline \hline
Path length & 0.3 - 1.0 & 0 - 15  & 104.58  $\pm$ 9.66 & 109.12 $\pm$ 11.32 & 4.25$\uparrow$  & 108.07   $\pm$ 9.91  & 3.29$\uparrow$  & 110.43  $\pm$ 11.87 & 5.45$\uparrow$  & 116.31  $\pm$ 16.83 & 10.63$\uparrow$  & 132.90      $\pm$ 30.43 & 23.85$\uparrow$  \\
$L$ [$m$] [$ \downarrow $]            &           & 15 - 30 & 104.58  $\pm$ 9.66 & 108.69 $\pm$ 11.59 & 3.86$\uparrow$  & 109.76   $\pm$ 10.78 & 4.84$\uparrow$  & 110.71  $\pm$ 12.16 & 5.70$\uparrow$  & 115.38  $\pm$ 16.27 & 9.82$\uparrow$   & 132.77      $\pm$ 31.25 & 23.75$\uparrow$  \\
            &           & 30 - 45 & 104.62  $\pm$ 9.73 & 109.82 $\pm$ 10.59 & 4.85$\uparrow$  & 110.24   $\pm$ 10.35 & 5.23$\uparrow$  & 111.06  $\pm$ 12.23 & 5.97$\uparrow$  & 116.30  $\pm$ 14.41 & 10.58$\uparrow$  & 132.44      $\pm$ 30.27 & 23.47$\uparrow$  \\
            &           & 45 - 50 & 104.62  $\pm$ 9.73 & 113.86 $\pm$ 11.00 & 8.46$\uparrow$  & 110.96   $\pm$ 10.36 & 5.88$\uparrow$  & 110.49  $\pm$ 11.92 & 5.46$\uparrow$  & 120.41  $\pm$ 13.47 & 14.03$\uparrow$  & 132.56      $\pm$ 30.92 & 23.56$\uparrow$  \\
            & 1.0 - 2.0 & 0 - 15  & 104.51  $\pm$ 9.68 & 107.12 $\pm$ 8.99  & 2.46$\uparrow$  & 106.54   $\pm$ 9.23  & 1.92$\uparrow$  & 107.07  $\pm$ 8.99  & 2.42$\uparrow$  & 110.50  $\pm$ 9.93  & 5.57$\uparrow$   & 132.65      $\pm$ 30.60 & 23.73$\uparrow$  \\
            &           & 15 - 30 & 104.45  $\pm$ 9.46 & 106.51 $\pm$ 9.11  & 1.95$\uparrow$  & 106.84   $\pm$ 9.18  & 2.27$\uparrow$  & 107.10  $\pm$ 8.97  & 2.51$\uparrow$  & 110.18  $\pm$ 9.59  & 5.35$\uparrow$   & 132.43      $\pm$ 30.33 & 23.63$\uparrow$  \\
            &           & 30 - 45 & 104.45  $\pm$ 9.53 & 107.78 $\pm$ 9.84  & 3.14$\uparrow$  & 108.59   $\pm$ 9.64  & 3.89$\uparrow$  & 107.07  $\pm$ 9.02  & 2.47$\uparrow$  & 112.57  $\pm$ 8.96  & 7.48$\uparrow$   & 132.40      $\pm$ 30.40 & 23.60$\uparrow$  \\
            &           & 45 - 50 & 104.38  $\pm$ 9.43 & 113.34 $\pm$ 10.48 & 8.23$\uparrow$  & 110.28   $\pm$ 9.91  & 5.50$\uparrow$  & 107.21  $\pm$ 9.21  & 2.67$\uparrow$  & 114.32  $\pm$ 9.32  & 9.09$\uparrow$   & 132.02      $\pm$ 30.30 & 23.39$\uparrow$  \\
            & 2.0 - 3.0 & 0 - 15  & 104.19  $\pm$ 9.71 & 107.10 $\pm$ 8.99  & 2.76$\uparrow$  & 105.99   $\pm$ 9.21  & 1.72$\uparrow$  & 107.41  $\pm$ 9.40  & 3.04$\uparrow$  & 110.57  $\pm$ 9.88  & 5.94$\uparrow$   & 132.32      $\pm$ 30.36 & 23.79$\uparrow$  \\
            &           & 15 - 30 & 104.21  $\pm$ 9.77 & 106.51 $\pm$ 9.46  & 2.18$\uparrow$  & 106.21   $\pm$ 9.13  & 1.90$\uparrow$  & 107.34  $\pm$ 9.31  & 2.96$\uparrow$  & 110.84  $\pm$ 9.27  & 6.16$\uparrow$   & 132.34      $\pm$ 30.33 & 23.78$\uparrow$  \\
            &           & 30 - 45 & 104.16  $\pm$ 9.69 & 106.91 $\pm$ 9.79  & 2.60$\uparrow$  & 108.44   $\pm$ 9.69  & 4.03$\uparrow$  & 107.38  $\pm$ 9.31  & 3.05$\uparrow$  & 112.51  $\pm$ 8.79  & 7.71$\uparrow$   & 132.69      $\pm$ 30.79 & 24.10$\uparrow$  \\
            &           & 45 - 50 & 104.17  $\pm$ 9.65 & 113.60 $\pm$ 10.56 & 8.67$\uparrow$  & 109.92   $\pm$ 9.97  & 5.38$\uparrow$  & 107.46  $\pm$ 9.40  & 3.11$\uparrow$  & 117.27  $\pm$ 10.27 & 11.83$\uparrow$  & 132.56      $\pm$ 29.94 & 23.98$\uparrow$  \\
            & 3.0 - 4.0 & 0 - 15  & 104.04  $\pm$ 9.51 & 106.71 $\pm$ 8.88  & 2.53$\uparrow$  & 105.34   $\pm$ 9.21  & 1.24$\uparrow$  & 107.57  $\pm$ 9.38  & 3.34$\uparrow$  & 111.53  $\pm$ 10.11 & 6.95$\uparrow$   & 133.16      $\pm$ 30.85 & 24.55$\uparrow$  \\
            &           & 15 - 30 & 104.06  $\pm$ 9.55 & 106.21 $\pm$ 9.26  & 2.04$\uparrow$  & 105.46   $\pm$ 9.27  & 1.33$\uparrow$  & 107.64  $\pm$ 9.56  & 3.38$\uparrow$  & 111.36  $\pm$ 9.30  & 6.77$\uparrow$   & 133.22      $\pm$ 30.97 & 24.58$\uparrow$  \\
            &           & 30 - 45 & 104.02  $\pm$ 9.48 & 106.73 $\pm$ 9.67  & 2.57$\uparrow$  & 107.99   $\pm$ 9.54  & 3.74$\uparrow$  & 107.37  $\pm$ 9.34  & 3.17$\uparrow$  & 112.43  $\pm$ 8.27  & 7.77$\uparrow$   & 131.89      $\pm$ 29.88 & 23.62$\uparrow$  \\
            &           & 45 - 50 & 104.16  $\pm$ 9.46 & 113.08 $\pm$ 10.39 & 8.22$\uparrow$  & 109.39   $\pm$ 9.64  & 4.90$\uparrow$  & 107.53  $\pm$ 9.38  & 3.19$\uparrow$  & 114.20  $\pm$ 8.70  & 9.20$\uparrow$   & 133.05      $\pm$ 30.42 & 24.36$\uparrow$  \\
            & 4.0 - 5.0 & 0 - 15  & 104.46  $\pm$ 9.62 & 107.29 $\pm$ 8.87  & 2.67$\uparrow$  & 105.20   $\pm$ 9.38  & 0.70$\uparrow$  & 108.68  $\pm$ 9.74  & 3.96$\uparrow$  & 111.72  $\pm$ 10.07 & 6.72$\uparrow$   & 133.52      $\pm$ 30.91 & 24.42$\uparrow$  \\
            &           & 15 - 30 & 104.50  $\pm$ 9.59 & 106.53 $\pm$ 9.28  & 1.92$\uparrow$  & 105.49   $\pm$ 9.52  & 0.94$\uparrow$  & 108.45  $\pm$ 9.45  & 3.71$\uparrow$  & 111.06  $\pm$ 9.22  & 6.08$\uparrow$   & 132.98      $\pm$ 30.62 & 23.98$\uparrow$  \\
            &           & 30 - 45 & 104.52  $\pm$ 9.72 & 106.96 $\pm$ 9.84  & 2.30$\uparrow$  & 107.60   $\pm$ 9.78  & 2.90$\uparrow$  & 108.87  $\pm$ 9.60  & 4.07$\uparrow$  & 112.82  $\pm$ 8.53  & 7.63$\uparrow$   & 133.27      $\pm$ 30.61 & 24.18$\uparrow$  \\
            &           & 45 - 50 & 104.56  $\pm$ 9.70 & 113.23 $\pm$ 10.76 & 7.97$\uparrow$  & 108.76   $\pm$ 9.53  & 3.94$\uparrow$  & 108.48  $\pm$ 9.70  & 3.69$\uparrow$  & 115.22  $\pm$ 9.29  & 9.71$\uparrow$   & 133.46      $\pm$ 30.25 & 24.29$\uparrow$  \\
\hline \hline
\end{tabular}
\caption{Path length (\ref{eqn:PathLength}) in meters denoted by $L$ for the A* planners and neural networks. Mean and standard deviation provided for the current‐informed A* path planner (Current), CFD informed, PatchGAN informed and 2D3DGAN informed A* planners. Also provided are the CFD\_NN and Current\_NN neural network approximations. Arrows indicate whether each method’s mean is higher (↑) or lower (↓) than the Current‐informed A* planner.}
\label{tab:Path_Length}
\end{sidewaystable}

\begin{sidewaystable}[!htb]
\centering
\scriptsize
\begin{tabular}{lll|l|ll|ll|ll|ll|ll}
\hline \hline
Metric       & Speed     & Angle   & Current            & CFD             & \%        & PatchGAN            & \%        & 2D3DGAN (SA)           & \%        & CFD\_NN            & \%         & Current\_NN            & \%         \\
\hline \hline
High V cells & 0.3 - 1.0 & 0 - 15  & 2.08    $\pm$ 3.77 & 1.71 $\pm$ 2.18 & 19.22$\downarrow$  & 3.75     $\pm$ 3.12 & 57.46$\uparrow$  & 1.34    $\pm$ 2.78 & 43.11$\downarrow$  & 4.58    $\pm$ 6.79 & 75.17$\uparrow$ & 2.74        $\pm$ 4.19 & 27.56$\uparrow$  \\
$N_{\text{high-velocity}}(P)$ [$n$] [$ \downarrow $]             
             &           & 15 - 30 & 5.61    $\pm$ 5.04 & 2.40 $\pm$ 1.44 & 57.19$\downarrow$  & 9.09     $\pm$ 5.34 & 62.06$\uparrow$  & 4.50    $\pm$ 3.43 & 19.82$\downarrow$  & 5.09    $\pm$ 4.86 & 9.23$\downarrow$  & 5.04        $\pm$ 4.45 & 10.15$\downarrow$ \\
             &           & 30 - 45 & 7.51    $\pm$ 5.89 & 2.64 $\pm$ 1.27 & 64.85$\downarrow$  & 7.74     $\pm$ 4.82 & 3.08$\uparrow$   & 6.39    $\pm$ 4.45 & 14.99$\downarrow$  & 5.31    $\pm$ 4.09 & 29.24$\downarrow$ & 7.15        $\pm$ 4.68 & 4.84$\downarrow$  \\
             &           & 45 - 50 & 8.40    $\pm$ 5.59 & 2.68 $\pm$ 1.17 & 68.10$\downarrow$  & 4.74     $\pm$ 2.92 & 43.57$\downarrow$ & 4.85    $\pm$ 3.47 & 42.26$\downarrow$ & 4.35    $\pm$ 3.69 & 48.21$\downarrow$ & 5.35        $\pm$ 3.10 & 36.31$\downarrow$ \\
             & 1.0 - 2.0 & 0 - 15  & 2.85    $\pm$ 4.78 & 2.63 $\pm$ 2.39 & 7.72$\downarrow$   & 6.30     $\pm$ 3.25 & 121.05$\uparrow$ & 2.28    $\pm$ 2.91 & 19.93$\downarrow$  & 4.70    $\pm$ 5.49 & 64.91$\uparrow$  & 3.15        $\pm$ 4.26 & 10.53$\uparrow$  \\
             &           & 15 - 30 & 7.34    $\pm$ 5.26 & 3.17 $\pm$ 1.38 & 56.80$\downarrow$  & 11.97    $\pm$ 4.97 & 63.20$\uparrow$  & 5.82    $\pm$ 3.82 & 20.66$\downarrow$  & 6.35    $\pm$ 4.99 & 13.47$\downarrow$ & 6.75        $\pm$ 4.92 & 8.03$\downarrow$  \\
             &           & 30 - 45 & 8.91    $\pm$ 5.65 & 3.24 $\pm$ 1.09 & 63.64$\downarrow$  & 8.72     $\pm$ 3.71 & 2.16$\uparrow$   & 7.83    $\pm$ 3.87 & 12.13$\downarrow$  & 5.51    $\pm$ 3.00 & 38.20$\downarrow$ & 8.92        $\pm$ 4.05 & 0.11$\uparrow$   \\
             &           & 45 - 50 & 9.20    $\pm$ 5.82 & 2.10 $\pm$ 1.03 & 77.83$\downarrow$  & 5.39     $\pm$ 2.79 & 41.51$\downarrow$ & 4.91    $\pm$ 2.83 & 46.67$\downarrow$ & 5.34    $\pm$ 3.71 & 41.96$\downarrow$ & 6.50        $\pm$ 2.83 & 29.35$\downarrow$ \\
             & 2.0 - 3.0 & 0 - 15  & 2.91    $\pm$ 3.90 & 2.13 $\pm$ 2.23 & 26.80$\downarrow$  & 5.34     $\pm$ 2.94 & 83.16$\uparrow$  & 2.48    $\pm$ 3.10 & 14.78$\downarrow$  & 4.18    $\pm$ 4.74 & 43.30$\uparrow$  & 3.14        $\pm$ 4.28 & 7.91$\uparrow$   \\
             &           & 15 - 30 & 7.88    $\pm$ 5.88 & 3.14 $\pm$ 1.49 & 60.16$\downarrow$  & 9.57     $\pm$ 4.58 & 21.38$\uparrow$  & 6.06    $\pm$ 3.99 & 23.11$\downarrow$  & 5.66    $\pm$ 4.42 & 28.16$\downarrow$ & 6.53        $\pm$ 5.21 & 17.14$\downarrow$ \\
             &           & 30 - 45 & 9.35    $\pm$ 5.44 & 3.53 $\pm$ 0.99 & 62.27$\downarrow$  & 8.45     $\pm$ 3.45 & 9.64$\downarrow$  & 8.29    $\pm$ 3.70 & 11.29$\downarrow$  & 5.29    $\pm$ 2.74 & 43.42$\downarrow$ & 9.02        $\pm$ 4.01 & 3.49$\downarrow$  \\
             &           & 45 - 50 & 8.28    $\pm$ 5.63 & 1.86 $\pm$ 1.32 & 77.54$\downarrow$  & 4.64     $\pm$ 2.35 & 43.98$\downarrow$ & 5.89    $\pm$ 2.41 & 28.84$\downarrow$  & 6.44    $\pm$ 4.33 & 22.15$\downarrow$ & 6.16        $\pm$ 3.32 & 25.63$\downarrow$ \\
             & 3.0 - 4.0 & 0 - 15  & 3.17    $\pm$ 4.95 & 2.18 $\pm$ 2.36 & 31.23$\downarrow$  & 4.61     $\pm$ 3.22 & 45.42$\uparrow$  & 2.55    $\pm$ 3.45 & 19.55$\downarrow$  & 4.29    $\pm$ 5.43 & 35.01$\uparrow$  & 2.87        $\pm$ 4.36 & 9.46$\uparrow$   \\
             &           & 15 - 30 & 7.58    $\pm$ 5.22 & 3.19 $\pm$ 1.48 & 57.93$\downarrow$  & 7.48     $\pm$ 3.85 & 1.36$\downarrow$  & 6.03    $\pm$ 3.76 & 20.42$\downarrow$  & 5.64    $\pm$ 4.88 & 25.60$\downarrow$ & 6.43        $\pm$ 4.90 & 15.17$\downarrow$ \\
             &           & 30 - 45 & 9.31    $\pm$ 5.56 & 3.86 $\pm$ 0.98 & 58.58$\downarrow$  & 7.88     $\pm$ 3.45 & 15.38$\downarrow$ & 8.92    $\pm$ 3.35 & 4.24$\downarrow$   & 5.64    $\pm$ 2.73 & 39.44$\downarrow$ & 9.38        $\pm$ 4.24 & 0.38$\downarrow$  \\
             &           & 45 - 50 & 8.76    $\pm$ 5.45 & 2.69 $\pm$ 1.12 & 69.30$\downarrow$  & 4.68     $\pm$ 2.46 & 46.56$\downarrow$ & 5.46    $\pm$ 2.28 & 37.66$\downarrow$  & 5.15    $\pm$ 2.80 & 41.20$\downarrow$ & 6.39        $\pm$ 3.25 & 27.05$\downarrow$ \\
             & 4.0 - 5.0 & 0 - 15  & 3.18    $\pm$ 3.90 & 2.38 $\pm$ 2.22 & 25.16$\downarrow$  & 3.76     $\pm$ 4.59 & 18.24$\uparrow$  & 2.78    $\pm$ 2.92 & 12.58$\downarrow$  & 4.86    $\pm$ 5.97 & 52.83$\uparrow$  & 3.02        $\pm$ 4.38 & 5.03$\uparrow$   \\
             &           & 15 - 30 & 7.24    $\pm$ 5.17 & 3.15 $\pm$ 1.30 & 56.50$\downarrow$  & 5.91     $\pm$ 2.91 & 18.41$\downarrow$  & 6.24    $\pm$ 3.99 & 13.78$\downarrow$  & 5.71    $\pm$ 4.67 & 21.09$\downarrow$ & 6.26        $\pm$ 5.05 & 13.55$\downarrow$ \\
             &           & 30 - 45 & 9.73    $\pm$ 5.74 & 3.85 $\pm$ 0.81 & 60.47$\downarrow$  & 6.37     $\pm$ 3.11 & 34.55$\downarrow$ & 9.57    $\pm$ 3.82 & 1.66$\downarrow$   & 5.61    $\pm$ 2.80 & 42.34$\downarrow$ & 9.36        $\pm$ 4.42 & 3.82$\downarrow$  \\
             &           & 45 - 50 & 8.91    $\pm$ 5.66 & 2.69 $\pm$ 0.87 & 69.82$\downarrow$  & 5.21     $\pm$ 3.16 & 41.56$\downarrow$ & 5.56    $\pm$ 2.71 & 37.58$\downarrow$  & 5.31    $\pm$ 3.33 & 40.41$\downarrow$ & 6.39        $\pm$ 3.32 & 28.32$\downarrow$ \\
\hline \hline
\end{tabular}
\caption{Number of high-velocity cells $N_{\text{high-velocity}}(P)$ (\ref{eqn:PathHighVelocity}) encountered on each trajectory $P$ for the A* planners and neural networks. Mean and standard deviation provided for the current‐informed A* path planner (Current), CFD informed, PatchGAN informed and 2D3DGAN informed A* planners. Also provided are the CFD\_NN and Current\_NN neural network approximations. Arrows indicate whether each method’s mean is higher (↑) or lower (↓) than the Current‐informed A* planner.}
\label{tab:High_Velocity_Cells}
\end{sidewaystable}

\begin{sidewaystable}[!htb]
\centering
\scriptsize
\begin{tabular}{lll|l|ll|ll|ll|ll|ll}
\hline \hline
Metric      & Speed     & Angle   & Current             & CFD               & \%           & PatchGAN             & \%           & 2D3DGAN (SA)            & \%           & CFD\_NN             & \%           & Current\_NN             & \%           \\
\hline \hline
Turb. Cells & 0.3 - 1.0 & 0 - 15  & 34.00   $\pm$ 17.73 & 58.00 $\pm$ 19.98 & 52.17$\uparrow$ & 52.00    $\pm$ 17.05 & 41.86$\uparrow$ & 45.00   $\pm$ 21.59 & 27.85$\uparrow$ & 55.00   $\pm$ 18.89 & 47.19$\uparrow$ & 42.00       $\pm$ 16.80 & 21.05$\uparrow$ \\
$N_{\text{turbulent}}(P)$ [$n$] [$ \downarrow $]
            &           & 15 - 30 & 39.00   $\pm$ 19.01 & 60.00 $\pm$ 19.79 & 42.42$\uparrow$ & 57.00    $\pm$ 19.08 & 37.50$\uparrow$ & 53.50   $\pm$ 20.08 & 31.35$\uparrow$ & 61.00   $\pm$ 19.17 & 44.00$\uparrow$ & 51.00       $\pm$ 16.23 & 26.67$\uparrow$ \\
            &           & 30 - 45 & 27.50   $\pm$ 20.88 & 42.00 $\pm$ 21.02 & 41.73$\uparrow$ & 43.50    $\pm$ 20.04 & 45.07$\uparrow$ & 43.00   $\pm$ 21.55 & 43.97$\uparrow$ & 41.00   $\pm$ 19.88 & 39.42$\uparrow$ & 38.00       $\pm$ 18.40 & 32.06$\uparrow$ \\
            &           & 45 - 50 & 24.00   $\pm$ 13.40 & 35.00 $\pm$ 18.28 & 37.29$\uparrow$ & 31.50    $\pm$ 15.77 & 27.03$\uparrow$ & 31.00   $\pm$ 14.61 & 25.45$\uparrow$ & 35.00   $\pm$ 17.44 & 37.29$\uparrow$ & 30.00       $\pm$ 15.13 & 22.22$\uparrow$ \\
            & 1.0 - 2.0 & 0 - 15  & 24.00   $\pm$ 9.82  & 44.00 $\pm$ 13.63 & 58.82$\uparrow$ & 42.00    $\pm$ 11.48 & 54.55$\uparrow$ & 36.00   $\pm$ 12.87 & 40.00$\uparrow$ & 44.00   $\pm$ 12.51 & 58.82$\uparrow$ & 35.00       $\pm$ 8.81  & 37.29$\uparrow$ \\
            &           & 15 - 30 & 25.00   $\pm$ 14.45 & 45.00 $\pm$ 14.17 & 57.14$\uparrow$ & 46.00    $\pm$ 12.99 & 59.15$\uparrow$ & 44.50   $\pm$ 13.49 & 56.12$\uparrow$ & 46.00   $\pm$ 13.72 & 59.15$\uparrow$ & 40.00       $\pm$ 11.93 & 46.15$\uparrow$ \\
            &           & 30 - 45 & 26.00   $\pm$ 14.27 & 36.00 $\pm$ 14.61 & 32.26$\uparrow$ & 38.00    $\pm$ 14.30 & 37.50$\uparrow$ & 39.00   $\pm$ 12.37 & 40.00$\uparrow$ & 35.00   $\pm$ 14.78 & 29.51$\uparrow$ & 36.00       $\pm$ 13.48 & 32.26$\uparrow$ \\
            &           & 45 - 50 & 27.00   $\pm$ 8.50  & 36.00 $\pm$ 14.21 & 28.57$\uparrow$ & 32.00    $\pm$ 11.59 & 16.95$\uparrow$ & 33.00   $\pm$ 8.00  & 20.00$\uparrow$ & 32.00   $\pm$ 14.81 & 16.95$\uparrow$ & 31.00       $\pm$ 10.48 & 13.79$\uparrow$ \\
            & 2.0 - 3.0 & 0 - 15  & 26.00   $\pm$ 11.88 & 44.00 $\pm$ 13.92 & 51.43$\uparrow$ & 44.00    $\pm$ 11.95 & 51.43$\uparrow$ & 37.00   $\pm$ 12.67 & 34.92$\uparrow$ & 46.00   $\pm$ 13.18 & 55.56$\uparrow$ & 36.00       $\pm$ 9.21  & 32.26$\uparrow$ \\
            &           & 15 - 30 & 30.00   $\pm$ 15.51 & 47.00 $\pm$ 13.69 & 44.16$\uparrow$ & 47.00    $\pm$ 12.35 & 44.16$\uparrow$ & 46.00   $\pm$ 12.96 & 42.11$\uparrow$ & 48.00   $\pm$ 13.80 & 46.15$\uparrow$ & 42.00       $\pm$ 12.08 & 33.33$\uparrow$ \\
            &           & 30 - 45 & 30.00   $\pm$ 13.72 & 38.00 $\pm$ 14.23 & 23.53$\uparrow$ & 41.00    $\pm$ 13.62 & 26.09$\uparrow$ & 43.00   $\pm$ 11.69 & 30.99$\uparrow$ & 37.00   $\pm$ 14.65 & 20.90$\uparrow$ & 38.00       $\pm$ 13.35 & 23.53$\uparrow$ \\
            &           & 45 - 50 & 28.00   $\pm$ 9.34  & 37.00 $\pm$ 14.79 & 27.69$\uparrow$ & 34.00    $\pm$ 11.53 & 19.35$\uparrow$ & 35.00   $\pm$ 8.62  & 22.22$\uparrow$ & 37.00   $\pm$ 15.79 & 27.69$\uparrow$ & 34.00       $\pm$ 11.58 & 19.35$\uparrow$ \\
            & 3.0 - 4.0 & 0 - 15  & 28.00   $\pm$ 11.44 & 46.00 $\pm$ 13.55 & 48.65$\uparrow$ & 45.00    $\pm$ 12.26 & 46.58$\uparrow$ & 38.00   $\pm$ 12.96 & 30.30$\uparrow$ & 46.00   $\pm$ 13.50 & 48.65$\uparrow$ & 37.00       $\pm$ 10.00 & 27.69$\uparrow$ \\
            &           & 15 - 30 & 34.00   $\pm$ 14.43 & 50.00 $\pm$ 13.13 & 38.10$\uparrow$ & 48.00    $\pm$ 12.63 & 34.15$\uparrow$ & 48.00   $\pm$ 12.62 & 34.15$\uparrow$ & 50.00   $\pm$ 13.34 & 38.10$\uparrow$ & 45.00       $\pm$ 11.81 & 27.85$\uparrow$ \\
            &           & 30 - 45 & 32.00   $\pm$ 14.17 & 40.00 $\pm$ 14.53 & 22.22$\uparrow$ & 41.00    $\pm$ 14.31 & 24.66$\uparrow$ & 43.00   $\pm$ 12.34 & 29.33$\uparrow$ & 39.00   $\pm$ 14.42 & 19.72$\uparrow$ & 40.00       $\pm$ 13.45 & 22.22$\uparrow$ \\
            &           & 45 - 50 & 30.00   $\pm$ 9.46  & 38.00 $\pm$ 14.24 & 23.53$\uparrow$ & 36.50    $\pm$ 11.55 & 19.55$\uparrow$ & 37.00   $\pm$ 9.75  & 20.90$\uparrow$ & 37.00   $\pm$ 13.48 & 20.90$\uparrow$ & 37.00       $\pm$ 11.54 & 20.90$\uparrow$ \\
            & 4.0 - 5.0 & 0 - 15  & 31.00   $\pm$ 12.31 & 47.00 $\pm$ 13.23 & 41.03$\uparrow$ & 37.00    $\pm$ 10.69 & 17.65$\uparrow$ & 41.00   $\pm$ 11.89 & 27.78$\uparrow$ & 48.00   $\pm$ 13.88 & 43.04$\uparrow$ & 39.00       $\pm$ 11.32 & 22.86$\uparrow$ \\
            &           & 15 - 30 & 33.00   $\pm$ 15.06 & 50.00 $\pm$ 13.21 & 40.96$\uparrow$ & 44.00    $\pm$ 14.21 & 28.57$\uparrow$ & 50.00   $\pm$ 12.42 & 40.96$\uparrow$ & 52.00   $\pm$ 13.29 & 44.71$\uparrow$ & 46.00       $\pm$ 12.07 & 32.91$\uparrow$ \\
            &           & 30 - 45 & 35.00   $\pm$ 14.17 & 40.00 $\pm$ 14.39 & 13.33$\uparrow$ & 42.00    $\pm$ 15.05 & 18.18$\uparrow$ & 45.00   $\pm$ 14.47 & 25.00$\uparrow$ & 40.50   $\pm$ 14.34 & 14.57$\uparrow$ & 41.00       $\pm$ 13.51 & 15.79$\uparrow$ \\
            &           & 45 - 50 & 33.00   $\pm$ 9.47  & 40.00 $\pm$ 14.11 & 19.18$\uparrow$ & 38.00    $\pm$ 11.69 & 14.08$\uparrow$ & 39.00   $\pm$ 11.41 & 16.67$\uparrow$ & 39.50   $\pm$ 13.37 & 17.93$\uparrow$ & 38.00       $\pm$ 11.58 & 14.08$\uparrow$ \\
\hline \hline
\end{tabular}
\caption{Number of turbulent cells $N_{\text{turbulent}}(P)$ (\ref{eqn:PathTurbulence}) encountered on each trajectory $P$ for the A* planners and neural networks. Mean and standard deviation provided for the current‐informed A* path planner (Current), CFD informed, PatchGAN informed and 2D3DGAN informed A* planners. Also provided are the CFD\_NN and Current\_NN neural network approximations. Arrows indicate whether each method’s mean is higher (↑) or lower (↓) than the Current‐informed A* planner.}
\label{tab:Turbulent_Cells}
\end{sidewaystable}

\begin{sidewaystable}[!htb]
\centering
\scriptsize
\begin{tabular}{lll|l|ll|ll|ll|ll|ll}
\hline \hline
Metric & Speed & Angle & Current & CFD & \% & PatchGAN & \% & 2D3DGAN (SA) & \% & CFD\_NN & \% & Current\_NN & \% \\
\hline \hline
Time & 0.3 - 1.0 & 0 - 15 & 67.32 $\pm$ 18.95 & 80.78 $\pm$ 597.65 & 18.18$\uparrow$ & 49.70 $\pm$ 28.48 & 30.12$\downarrow$ & 45.15 $\pm$ 73.90 & 39.42$\downarrow$ & 2.90E-04 $\pm$ 1.60E-04 & & 3.40E-04 $\pm$ 5.31E-02 & \\
$T$ [$s$] [$ \downarrow $] & & 15 - 30 & 67.47 $\pm$ 17.74 & 80.79 $\pm$ 222.33 & 17.96$\uparrow$ & 44.97 $\pm$ 31.45 & 40.04$\downarrow$ & 43.62 $\pm$ 55.40 & 42.95$\downarrow$ & 3.00E-04 $\pm$ 3.60E-04 & & 3.60E-04 $\pm$ 3.60E-04 & \\
 & & 30 - 45 & 69.11 $\pm$ 18.00 & 45.49 $\pm$ 34.97 & 41.22$\downarrow$ & 35.66 $\pm$ 24.48 & 63.85$\downarrow$ & 42.77 $\pm$ 58.18 & 47.08$\downarrow$ & 2.80E-04 $\pm$ 1.40E-04 & & 3.30E-04 $\pm$ 1.20E-04 & \\
 & & 45 - 50 & 68.79 $\pm$ 17.97 & 41.80 $\pm$ 25.79 & 48.81$\downarrow$ & 30.74 $\pm$ 19.86 & 76.45$\downarrow$ & 44.06 $\pm$ 42.37 & 43.83$\downarrow$ & 2.80E-04 $\pm$ 1.50E-04 & & 3.30E-04 $\pm$ 1.30E-04 & \\
 & 1.0 - 2.0 & 0 - 15 & 68.84 $\pm$ 17.48 & 60.92 $\pm$ 26.60 & 12.20$\downarrow$ & 47.73 $\pm$ 17.47 & 36.22$\downarrow$ & 50.16 $\pm$ 17.22 & 31.39$\downarrow$ & 2.90E-04 $\pm$ 1.80E-04 & & 3.50E-04 $\pm$ 2.00E-04 & \\
 & & 15 - 30 & 68.82 $\pm$ 17.47 & 64.38 $\pm$ 22.33 & 6.66$\downarrow$ & 47.28 $\pm$ 17.04 & 37.12$\downarrow$ & 50.24 $\pm$ 17.10 & 31.21$\downarrow$ & 2.90E-04 $\pm$ 2.40E-04 & & 3.40E-04 $\pm$ 2.20E-04 & \\
 & & 30 - 45 & 68.40 $\pm$ 17.42 & 52.51 $\pm$ 24.41 & 26.29$\downarrow$ & 38.98 $\pm$ 18.53 & 54.79$\downarrow$ & 50.65 $\pm$ 17.00 & 29.82$\downarrow$ & 2.90E-04 $\pm$ 2.20E-04 & & 3.40E-04 $\pm$ 2.00E-04 & \\
 & & 45 - 50 & 68.44 $\pm$ 17.21 & 45.66 $\pm$ 23.73 & 39.93$\downarrow$ & 34.80 $\pm$ 17.18 & 65.17$\downarrow$ & 50.17 $\pm$ 17.36 & 30.81$\downarrow$ & 3.00E-04 $\pm$ 4.00E-04 & & 3.70E-04 $\pm$ 3.90E-04 & \\
 & 2.0 - 3.0 & 0 - 15 & 68.92 $\pm$ 18.60 & 53.44 $\pm$ 24.20 & 25.31$\downarrow$ & 54.28 $\pm$ 18.12 & 23.77$\downarrow$ & 49.49 $\pm$ 17.25 & 32.83$\downarrow$ & 3.10E-04 $\pm$ 4.90E-04 & & 3.80E-04 $\pm$ 4.50E-04 & \\
 & & 15 - 30 & 68.86 $\pm$ 18.61 & 62.34 $\pm$ 22.46 & 9.95$\downarrow$ & 53.29 $\pm$ 17.52 & 25.50$\downarrow$ & 49.26 $\pm$ 17.22 & 33.19$\downarrow$ & 3.00E-04 $\pm$ 3.80E-04 & & 3.60E-04 $\pm$ 3.70E-04 & \\
 & & 30 - 45 & 68.84 $\pm$ 18.14 & 55.87 $\pm$ 24.14 & 20.79$\downarrow$ & 46.01 $\pm$ 20.09 & 39.75$\downarrow$ & 49.00 $\pm$ 17.20 & 33.68$\downarrow$ & 3.10E-04 $\pm$ 4.50E-04 & & 3.70E-04 $\pm$ 4.50E-04 & \\
 & & 45 - 50 & 68.75 $\pm$ 18.06 & 47.19 $\pm$ 24.30 & 37.20$\downarrow$ & 41.45 $\pm$ 19.47 & 49.55$\downarrow$ & 48.51 $\pm$ 17.41 & 34.53$\downarrow$ & 2.80E-04 $\pm$ 1.70E-04 & & 3.30E-04 $\pm$ 1.50E-04 & \\
 & 3.0 - 4.0 & 0 - 15 & 68.49 $\pm$ 17.98 & 51.18 $\pm$ 24.56 & 28.92$\downarrow$ & 56.89 $\pm$ 17.89 & 18.51$\downarrow$ & 43.64 $\pm$ 17.13 & 44.32$\downarrow$ & 2.90E-04 $\pm$ 2.50E-04 & & 3.40E-04 $\pm$ 4.90E-04 & \\
 & & 15 - 30 & 69.45 $\pm$ 18.15 & 61.61 $\pm$ 22.16 & 11.97$\downarrow$ & 56.14 $\pm$ 18.08 & 21.19$\downarrow$ & 44.51 $\pm$ 17.26 & 43.76$\downarrow$ & 2.90E-04 $\pm$ 2.90E-04 & & 3.50E-04 $\pm$ 2.80E-04 & \\
 & & 30 - 45 & 68.38 $\pm$ 18.13 & 53.80 $\pm$ 24.08 & 23.87$\downarrow$ & 46.19 $\pm$ 21.06 & 38.73$\downarrow$ & 44.42 $\pm$ 17.12 & 42.47$\downarrow$ & 2.80E-04 $\pm$ 1.50E-04 & & 3.30E-04 $\pm$ 1.30E-04 & \\
 & & 45 - 50 & 68.16 $\pm$ 17.89 & 46.29 $\pm$ 24.43 & 38.23$\downarrow$ & 42.51 $\pm$ 21.11 & 46.35$\downarrow$ & 44.29 $\pm$ 17.08 & 42.46$\downarrow$ & 2.90E-04 $\pm$ 2.20E-04 & & 3.40E-04 $\pm$ 2.00E-04 & \\
 & 4.0 - 5.0 & 0 - 15 & 68.75 $\pm$ 17.95 & 52.77 $\pm$ 24.51 & 26.31$\downarrow$ & 48.77 $\pm$ 15.83 & 34.00$\downarrow$ & 40.59 $\pm$ 15.87 & 51.52$\downarrow$ & 3.10E-04 $\pm$ 2.30E-04 & & 3.50E-04 $\pm$ 2.60E-04 & \\
 & & 15 - 30 & 66.98 $\pm$ 19.00 & 61.75 $\pm$ 21.93 & 8.11$\downarrow$ & 46.95 $\pm$ 16.72 & 35.15$\downarrow$ & 40.17 $\pm$ 15.70 & 50.03$\downarrow$ & 3.00E-04 $\pm$ 3.20E-04 & & 3.60E-04 $\pm$ 3.10E-04 & \\
 & & 30 - 45 & 65.23 $\pm$ 19.95 & 53.09 $\pm$ 24.36 & 20.52$\downarrow$ & 39.25 $\pm$ 18.61 & 49.73$\downarrow$ & 38.97 $\pm$ 15.67 & 50.40$\downarrow$ & 2.80E-04 $\pm$ 1.50E-04 & & 3.30E-04 $\pm$ 1.30E-04 & \\
 & & 45 - 50 & 65.51 $\pm$ 19.83 & 43.27 $\pm$ 25.10 & 40.91$\downarrow$ & 37.36 $\pm$ 18.03 & 54.74$\downarrow$ & 38.69 $\pm$ 16.16 & 51.48$\downarrow$ & 2.90E-04 $\pm$ 2.10E-04 & & 3.40E-04 $\pm$ 1.90E-04 & \\
\hline \hline
\end{tabular}
\caption{Compute time in seconds, denoted by $T$, for the A* planners and neural networks. Mean and standard deviation provided for the current‐informed A* path planner (Current), CFD informed, PatchGAN informed and 2D3DGAN informed A* planners. Also provided are the CFD\_NN and Current\_NN neural network approximations. Arrows indicate whether each method’s mean is higher (↑) or lower (↓) than the Current‐informed A* planner.}
\label{tab:Compute_Time}
\end{sidewaystable}

\end{document}